%% file: iclr2026_conference.tex
\documentclass{article} 
\usepackage{iclr2026_conference}
\usepackage{times}
\input{math_commands.tex}

\usepackage{natbib}   
\setcitestyle{numbers,comma,square,sort}

\usepackage{hyperref}
\usepackage{url}
\usepackage{comment}
\usepackage{graphicx}
\usepackage[table]{xcolor}
\usepackage[utf8]{inputenc} 
\usepackage[T1]{fontenc}    
\usepackage{hyperref}       
\usepackage{url}            
\usepackage{amsfonts}       
\usepackage{nicefrac}       
\usepackage{microtype}      
\usepackage{xcolor}         
\usepackage{enumerate}
\usepackage{enumitem}
\usepackage{subcaption}
\usepackage{mathtools}
\usepackage{array}
\usepackage{multirow}
\usepackage{siunitx}
\usepackage{caption}
\usepackage{placeins}
\usepackage{algorithmic}
\usepackage{pifont}
\usepackage{threeparttable}
\usepackage{adjustbox}
\usepackage{booktabs}
\usepackage{tikz}
\usepackage{physics}  
\usepackage{mathtools}
\usepackage{amsmath, amssymb, amsthm, graphicx}
\usepackage{algorithm, algorithmic}
\usepackage{longtable}
\usepackage{bm}
\usepackage[most]{tcolorbox}
\usepackage{makecell}
\usepackage{tabularx}
\usepackage{xcolor} 
\usepackage{pifont} 

\newcommand{\PromptEncoder}{\operatorname*{PromptEncoder}}
\newcolumntype{Y}{>{\centering\arraybackslash}X}
\definecolor{headbg}{RGB}{245,245,248}
\definecolor{motcol}{RGB}{230,245,234}
\definecolor{yesbg}{RGB}{222,245,225}
\definecolor{nobg}{RGB}{250,229,229}
\newcommand{\Yes}{\cellcolor{yesbg}\(\checkmark\)}
\newcommand{\No}{\cellcolor{nobg}\(\times\)}

\definecolor{mycolor}{HTML}{4779c4}
\definecolor{lb}{HTML}{E8F4FF}
\definecolor{db}{HTML}{4A90E2}

\definecolor{Header}{RGB}{245,245,245}   
\definecolor{Accent}{RGB}{230,249,235}   
\definecolor{headergray}{gray}{0.9}
\definecolor{fgtmgreen}{RGB}{225,255,225}

\definecolor{headerbg}{HTML}{C7DBF4}      
\definecolor{altrow}{HTML}{F3F6FB}        
\definecolor{highlightrow}{HTML}{FFF4D7}  

\definecolor{IDblue}{RGB}{0,102,204}
\definecolor{OODgreen}{RGB}{0,170,68}

\definecolor{lightgray}{RGB}{240,240,240}
\definecolor{lightblue}{RGB}{220,230,250}
\definecolor{lightgreen}{RGB}{225,245,225}

\definecolor{Hdr}{RGB}{245,247,250}    
\definecolor{Acc}{RGB}{30,64,175}      
\definecolor{AvgBG}{RGB}{250,252,255}  
\definecolor{TimeBG}{RGB}{250,252,250} 
\definecolor{Top}{RGB}{226,246,228}    
\definecolor{Fast}{RGB}{224,236,248}   
\definecolor{Block}{RGB}{248,248,248}  

\newcolumntype{A}{>{\columncolor{AvgBG}}c}   
\newcolumntype{T}{>{\columncolor{TimeBG}}c}  

\newcommand{\best}[1]{\cellcolor{Top}\textbf{#1}}
\newcommand{\fast}[1]{\cellcolor{Fast}\textbf{#1}}
\newcommand{\topaccent}{\arrayrulecolor{Acc}\specialrule{1pt}{0pt}{4pt}\arrayrulecolor{black}}
\newcommand{\bottomaccent}{\arrayrulecolor{Acc}\specialrule{1pt}{4pt}{0pt}\arrayrulecolor{black}}
\newcommand{\tighttable}{\small\setlength{\tabcolsep}{6pt}\renewcommand{\arraystretch}{1.12}}

\iclrfinalcopy 

\title{
Mixture of Thoughts: Learning to Aggregate What Experts Think, Not Just What They Say
}

\author{
Jacob-Fein Ashley$^{1}$, 
Dhruv Parikh$^{1}$, 
Rajgopal Kannan$^{2}$, 
Viktor Prasanna$^{1}$ \\
$^{1}$University of Southern California \quad 
$^{2}$DEVCOM ARL Army Research Office \\
\texttt{\{feinash,dhruvash,prasanna\}@usc.edu},\;
\texttt{rajgopal.kannan.civ@army.mil}
}

\begin{document}
\maketitle

\input{files/abstract.tex}

\input{files/intro.tex}

\input{files/related.tex}
\input{files/method.tex}
\input{files/experiments.tex}

\input{files/conclusion.tex}

\newpage

\bibliographystyle{abbrv}      
\bibliography{references}

\newpage
\appendix

\input{files/appendix_new}

\end{document}

%% file: math_commands.tex
\usepackage{amsmath,amsfonts,bm}

\def\eqref#1{equation~\ref{#1}}

\def\1{\bm{1}}

\DeclareMathAlphabet{\mathsfit}{\encodingdefault}{\sfdefault}{m}{sl}
\SetMathAlphabet{\mathsfit}{bold}{\encodingdefault}{\sfdefault}{bx}{n}



%% file: files/abstract.tex

\begin{abstract}
Open-source Large Language Models (LLMs) increasingly specialize by domain (e.g., math, code, general reasoning), motivating systems that leverage complementary strengths across models. Prior multi-LLM approaches either (i) \emph{route} a query to one or a few experts and generate independently, (ii) \emph{aggregate} outputs from each model via costly multi-turn exchanges, or (iii) \emph{fuse} weights into a single model—typically requiring architectural homogeneity. We introduce \emph{Mixture of Thoughts} (MoT), a simple method for \emph{latent-level collaboration} among heterogeneous experts under a global routing scheme. For each query, a lightweight router selects top-$K$ experts and designates a primary expert; uniformly placed \emph{interaction layers} project hidden states into a shared latent space where the primary expert performs cross-attention over its active (selected) peers. Pre-trained experts remain frozen; only the router and the lightweight interaction layers are trained with a novel joint training objective that improves both the expert selection and inter-expert collaboration. Across five in-distribution (ID) and three out-of-distribution (OOD) benchmarks, MoT surpasses the current routing and aggregation-based state-of-the-art, \textsc{Avengers}, by +0.38\% and +2.92\%, respectively. Further, MoT significantly outperforms the best-performing single model. It achieves this with single-pass inference, runtime comparable to routing baselines, and none of the overheads of iterative aggregation. MoT offers a simple latent-space mechanism for combining heterogeneous LLMs, a practical step toward broader multi-LLM collaboration. Our code is publicly available at \url{https://github.com/jacobfa/mot}.
\end{abstract}


\vspace{-10pt}

\begin{figure}[h]
    \centering
    \includegraphics[width=0.65\linewidth]{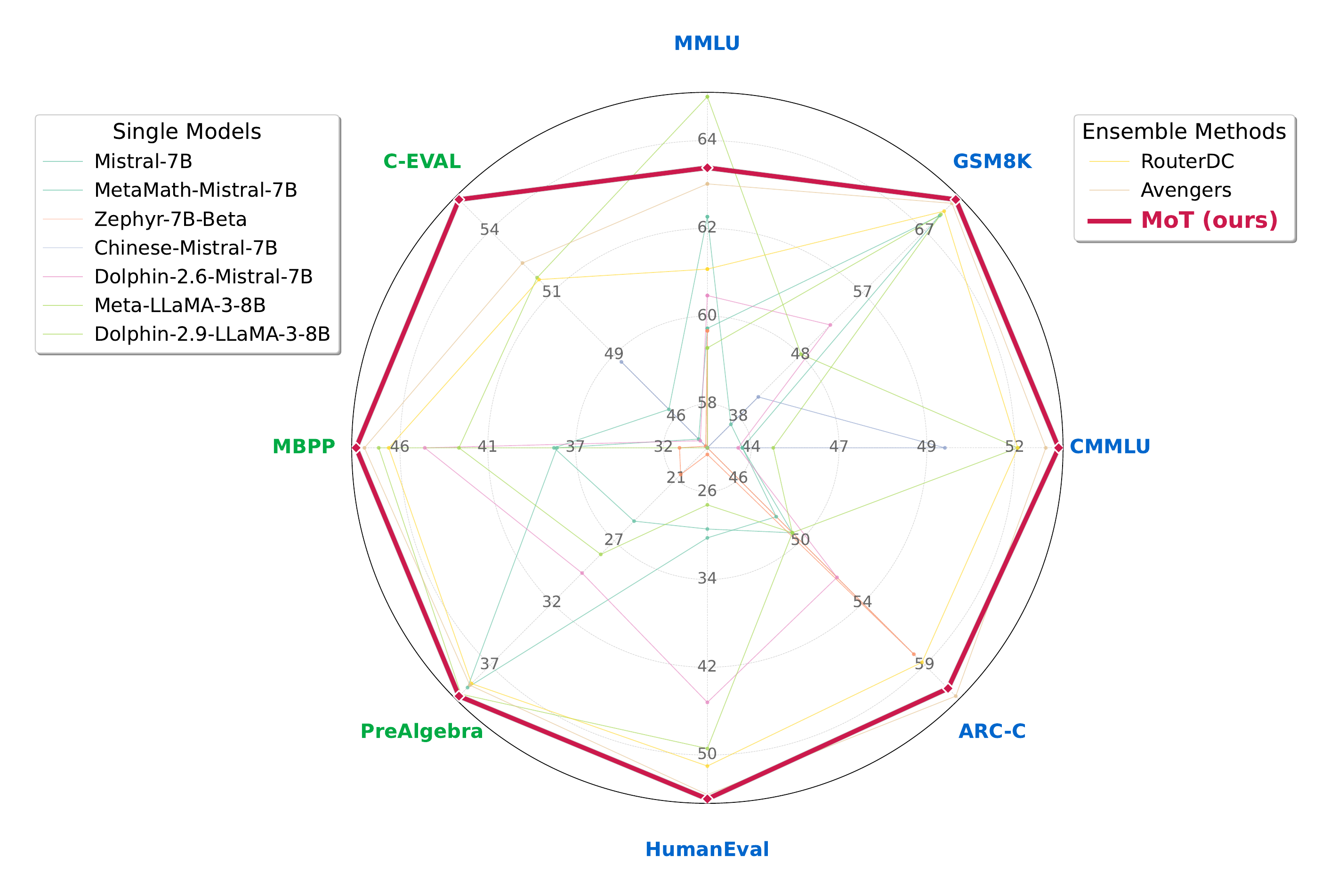}
    \caption{Accuracy radar plot across \textcolor{OODgreen}{OOD} and 
         \textcolor{IDblue}{ID} tasks, showing MoT 
        outperforming base models and state-of-the-art routing baselines. Scores are normalized to a $0-100$ range for visualization.}
    \label{fig:radar}
\end{figure}

%% file: files/intro.tex
\section{Introduction}
\label{sec:intro}

Large Language Models (LLMs) have reshaped applications in mathematics, coding, multimodal reasoning, and language understanding~\cite{llmsurvey}. Open-source LLMs (e.g., Mistral ~\cite{mistral7b}, LLaMA~3~\cite{llama-3}, LLaMA~4~\cite{llama-4}, Qwen~3\mbox{~\cite{qwen-3})} are competitive, yet closed-source systems (GPT-5~\cite{gpt-5}, Claude Opus~4.1~\cite{claude-opus-4.1}, Grok~4~\cite{grok-4}, Gemini~2.5~\cite{gemini-2.5}) still lead on widely used and highly competitive benchmarks, including MMLU~\cite{mmlu}, GSM8K~\cite{gsm8k}, BBH~\cite{big-bench}, HumanEval~\cite{humaneval}, ARC-Challenge~\cite{ARC-Challenge}, GPQA~\cite{GPQA}, MMMU/MMMU-Pro~\cite{mmmu,mmmu-pro}, SWE-Bench Verified~\cite{swe-bench,swe-bench-verified}, and Humanity's Last Exam~\cite{humanity-last-exam}.

Specialized open-source models excel in particular skills—math~\cite{meta-math}, code~\cite{llm-coding}, or commonsense~\cite{llm-common-sense-reasoning}—suggesting complementary expertise that could be combined. Prior multi-LLM efforts do so via: (i) \emph{selective routing} to one or a few experts~\cite{routerdc, zooter, embed-llm, routellm, graphrouter, routerbench, universal-routing, llm-blender, model-sat, avengers, deepen}, sometimes with ensemble sampling~\cite{avengers}; (ii) \emph{response-level collaboration} (mixture-of-agents)~\cite{moa, sparse-moa, self-moa, symbolic-moe}; and (iii) \emph{parameter-space fusion} (merging, continual learning, distillation)~\cite{ties-merging, DARE, model-swarms, lorahub, genome, model-fusion-homogeneity, fusellm}. These strategies are coarse-grained and/or compute-heavy, often preserving only narrow skills and weakening generalization~\cite{survey-multiple-llm}.

\noindent\textbf{Limitations.}
Learned routing (e.g., \textsc{RouterDC}~\cite{routerdc}) maps each query to a single expert, binding performance tightly to the chosen expert without any cross-model interaction. Multi-expert routing (e.g., \textsc{Avengers}~\cite{avengers}, current state of the art) relaxes this by selecting multiple experts, but combines their outputs through ensembling via sampling and voting \cite{wang2022self, multi-sample}, rather than true collaboration. While response-level collaboration (e.g., \textsc{MoA}~\cite{moa}) aggregates final outputs (responses) from multiple models, it does so at the cost of multi-turn iterative exchanges between models. Parameter fusion approaches (e.g., \textsc{TIES-Merging}~\cite{ties-merging}, \textsc{FuseLLM}~\cite{fusellm}) require architectural homogeneity and collapse specialization into a single set of weights, removing per-query adaptivity.

\begin{figure}[!h]
    \centering
    \includegraphics[width=0.95\linewidth]{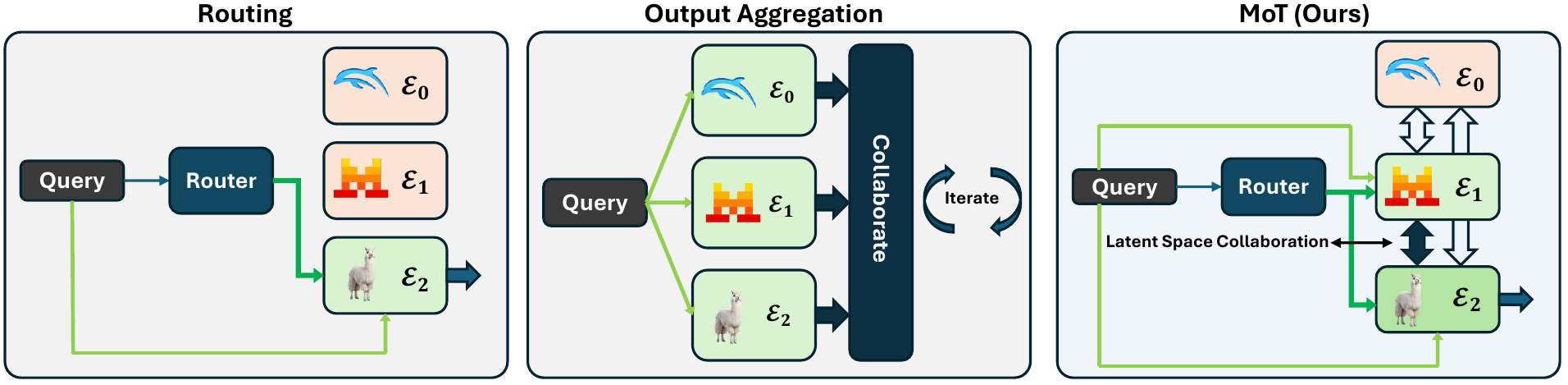}
    \caption{Prior paradigms (routing, output aggregation) vs.\ our \textsc{MoT} schematic. Parameter fusion schematic not shown as it does not support heterogeneous models. Dark green lines are for selected experts.}
    \label{fig:comparison}
\end{figure}

\noindent\textbf{Mixture of Thoughts (MoT).}
All prior paradigms lack fine-grained interaction in the latent space, restricting collaboration to outputs, weight fusion, or naive ensembling. To address this gap, we propose \emph{Mixture of Thoughts} (\textsc{MoT}), a method for \emph{latent-level collaboration} among heterogeneous experts under a global router. For each query, a lightweight trainable router selects the top-$K$ experts and designates a \emph{primary} expert. Within each expert, we uniformly place \emph{interaction layers} composed of lightweight adapters (projections) and a cross-attention module. At these layers, all active experts' hidden states are projected into a shared latent space and integrated into the primary expert via cross-attention—enabling fine-grained collaboration while keeping each expert's backbone frozen. A joint training objective optimizes both the router and interaction layers. \textsc{MoT} performs a \emph{single forward pass} through selected experts, achieving routing-like efficiency with richer representational capacity than single- and multi-model routing, response-level aggregation, and parameter fusion, while supporting fully heterogeneous experts with dynamic, per-query adaptive collaboration (Fig. \ref{fig:comparison}).

Unlike prior work that aggregates only at the output (\emph{‘say’}), \textsc{MoT} enables experts to integrate hidden representations (\emph{‘thoughts’}) via interaction layers.

\noindent
We evaluate \textsc{MoT} on language understanding, code generation, and mathematical reasoning under in-distribution (ID) and out-of-distribution (OOD) settings. \textsc{MoT} attains state-of-the-art performance among open-source multi-LLM methods: it surpasses the strongest single model (Dolphin-2.9-LLaMA-3-8B) by \textbf{+10.95}\% (ID) and \textbf{+9.03}\% (OOD), improves over \textsc{RouterDC} by \textbf{+3.40}\% (ID) and \textbf{+4.52}\% (OOD), and exceeds \textsc{Avengers} by \textbf{+0.38}\% (ID) and \textbf{+2.92}\% (OOD), with comparable wall-clock cost (51.8 vs.\ 51.3 minutes). \textsc{MoT} is robust to the number of selected experts, expert drop-out, the number of interaction layers (and their placement), and shared latent-space dimensionality.

{\scriptsize
\setlength{\tabcolsep}{6pt}%
\renewcommand{\arraystretch}{1.12}
\noindent
\begin{tabularx}{\linewidth}{>{\raggedright\arraybackslash}X *{4}{Y}}
\rowcolor{headbg}
\textbf{Capability} & \textbf{R} & \textbf{C} & \textbf{F} & \cellcolor{motcol}\textbf{MoT} \\
\midrule
Latent-level collaboration         & \No  & \No  & \No  & \Yes \\
Cost-efficient (single-pass)       & \Yes & \No  & \Yes & \Yes \\
Supports heterogeneous experts     & \Yes & \Yes & \No  & \Yes \\
Per-query collaborative adaptation & \No  & \Yes & \No  & \Yes \\
Beyond output-level fusion         & \No  & \No  & \Yes & \Yes \\
\bottomrule
\end{tabularx}

\vspace{0.3ex}
\noindent
\textbf{Representative works:}
\textbf{R} (routing):\ \cite{routerdc, zooter, embed-llm, routellm, universal-routing, llm-blender, graphrouter, routerbench, model-sat, avengers, deepen};\;
\textbf{C} (response-level collaboration):\ \cite{moa, sparse-moa, self-moa, symbolic-moe, avengers};\;
\textbf{F} (fusion):\ \cite{ties-merging, DARE, genome, model-fusion-homogeneity, model-swarms, lorahub, fusellm};\;
}

\vspace{-0.25em}
\noindent
\textbf{Contributions.}
(i) \textsc{MoT}: combines a lightweight global router with \emph{latent-level} interaction layers, enabling per-query collaboration among heterogeneous, frozen experts. 
(ii) A joint training objective optimizes the router and interaction layer parameters.
(iii) \emph{Single-pass} multi-expert inference yields lower overhead than response-level collaborative methods (e.g., \textsc{MoA}) and greater representational capacity than routing-only baselines (e.g., \textsc{RouterDC}, \textsc{Avengers}). 
(iv) Consistent ID/OOD gains: \textbf{+3.40}\% / \textbf{+4.52}\% over \textsc{RouterDC}, and \textbf{+0.38}\% / \textbf{+2.92}\% over \textsc{Avengers}, at comparable wall-clock cost. 
(v) Robust to expert count, drop-out, interaction depth/placement, and shared latent size; preserves specialization while enabling dynamic query-level collaboration.

%% file: files/related.tex
\section{Related Work}
\label{sec:related}

\textbf{Routing (Selection-Based).}
Routing-based integration methods use a learned or heuristic router to select one (or a small subset) of experts per query. \textsc{RouterDC}~\cite{routerdc} trains a parameter-efficient dual-contrastive router; \textsc{Zooter}~\cite{zooter} employs zero-shot prompt scoring; \textsc{EmbedLLM}~\cite{embed-llm} embeds queries and models in a joint space; \textsc{RouteLLM}~\cite{routellm} learns task-specific gates; and \textsc{GraphRouter}~\cite{graphrouter} models skill relationships over a graph. \textsc{RouterBench}~\cite{routerbench} evaluates routers across tasks, and~\cite{universal-routing} proposes task-agnostic, transferable routers. Multi-model variants (\textsc{LLMBlender}~\cite{llm-blender}, \textsc{Model-SAT}~\cite{model-sat}, \textsc{Avengers}~\cite{avengers}, \textsc{Deepen}~\cite{deepen}) select top-$K$ experts and combine outputs via ensembling (sampling \& voting, \cite{multi-sample,wang2022self}). These approaches typically fuse only \emph{final} generations, discarding intermediate states and making performance tightly depend on the routed experts’ outputs. In contrast, \textsc{MoT} preserves the efficiency of selection while enabling \emph{latent}-level exchange among the top-$K$ routed experts, allowing the primary expert to efficiently  aggregate token-level information across experts, in a single-pass.

\textbf{Response-Level Collaboration.}
Agentic systems coordinate multiple LLMs to iteratively refine responses: \textsc{MoA}~\cite{moa} orchestrates role-specialized agents, \textsc{Sparse-MoA}~\cite{sparse-moa} activates a subset per turn, \textsc{Self-MoA}~\cite{self-moa} adds self-reflection, and \textsc{Symbolic-MoE}~\cite{symbolic-moe} blends symbolic reasoning. While effective, these methods incur high inference cost from multi-round generation, interacting only at the response level. \textsc{MoT} attains collaborative benefits without iterative turns by sharing intermediate representations: active experts contribute projected hidden states that the primary expert cross-attends to within interaction layers, improving generation quality at single-pass inference cost.

\textbf{Parameter Fusion / Distillation.}
Weight-space integration merges experts into a single model, often assuming architectural homogeneity: \textsc{TIES-Merging}~\cite{ties-merging} and \textsc{DARE}~\cite{DARE} merge parameters; \textsc{Model-Swarms}~\cite{model-swarms}, \textsc{LoRAHub}~\cite{lorahub}, and \textsc{Genome}~\cite{genome} compose specialists via incremental or evolutionary search; \cite{model-fusion-homogeneity} discusses homogeneity and scalability constraints of these approaches. Distillation methods such as \textsc{FuseLLM}~\cite{fusellm} transfer multi-expert knowledge via generation supervision. These yield a single efficient model but collapse collaboration into static weights, diluting specialization and removing per-query adaptivity. \textsc{MoT} instead keeps all backbones frozen and heterogeneous, adds only lightweight projectors and cross-attention, and adapts collaboration per query through global routing and token-level latent integration.

\textbf{Our work (MoT).}
Prior work has centered on either routing queries to the most suitable expert, aggregating outputs through costly multi-turn interactions, or collapsing models via fusion and distillation that require architectural homogeneity and sacrifice diversity. In contrast, \textsc{MoT} introduces latent-space collaboration across heterogeneous experts, retaining individual model specialization while enabling efficient single-pass inference.

%% file: files/method.tex
\section{Method}
\label{sec:arch}

We now describe the proposed \emph{Mixture of Thoughts} (MoT) framework. Section~\ref{sec:overview} provides a high-level overview of the architecture. 
Section~\ref{sec:routing} details the global routing mechanism that selects a sparse subset of expert models from the given pool of LLMs. 
Section~\ref{sec:interaction} introduces the interaction layers that enable latent-space collaboration among routed experts. 
Section~\ref{sec:training} summarizes the training strategy, and Section~\ref{sec:inference} describes the inference procedure.

\subsection{Overview}
\label{sec:overview}

MoT integrates multiple pre-trained decoder-only language models to enable collaborative reasoning without modifying their backbone parameters. 
The system maintains $M$ frozen experts $\{\mathcal{E}_0, \ldots, \mathcal{E}_{M-1}\}$, which may differ in size, architecture, training corpus, or domain specialization. 
For each input, a lightweight global router activates a sparse Top-$K$ subset of experts, designating the highest-scoring one as the \emph{primary expert} responsible for output generation. 
Collaboration occurs through lightweight interaction layers that operate in a shared latent space, allowing the primary expert to integrate hidden representations from its peers via cross-attention. 
Unlike multi-turn approaches (e.g., \cite{moa, avengers}), MoT requires only a single generation pass, offering both efficiency and fine-grained inter-expert collaboration.

\begin{figure}[!htbp]
    \centering
    \includegraphics[width=0.92\linewidth]{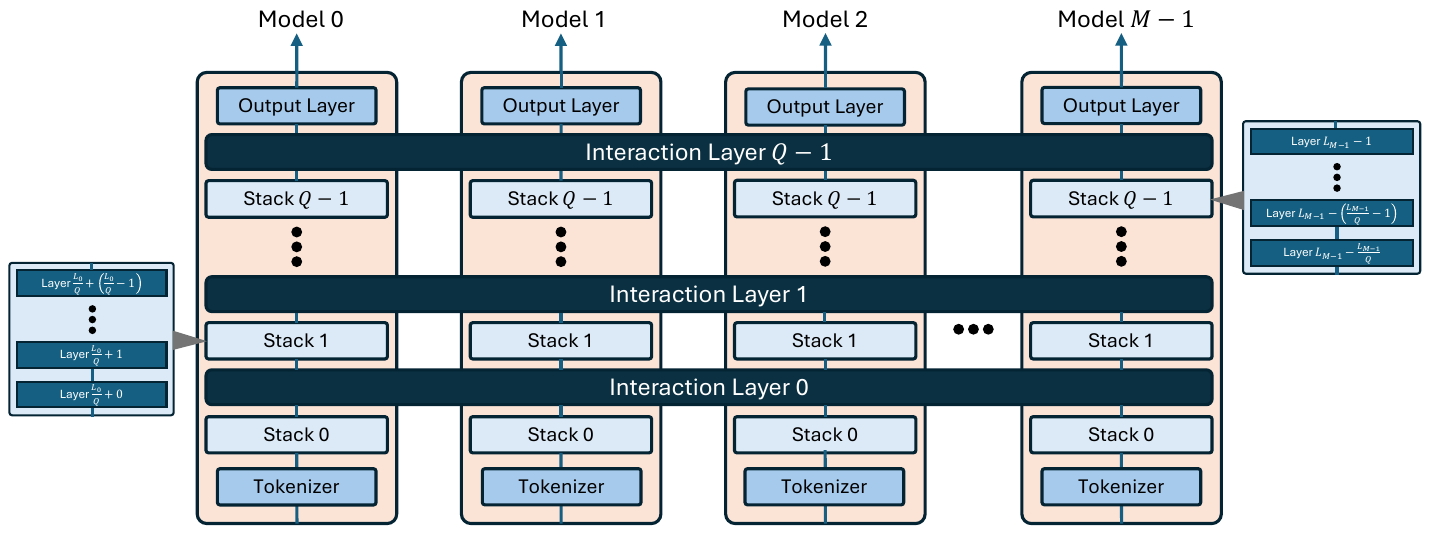}
    \caption{\textbf{Heterogeneous expert integration in MoT.} 
    Each expert with $L_m$ layers is divided into $Q$ stacks, enabling $Q$ levels of interaction. 
    After each stack, routed experts project hidden states into a shared latent space; the primary expert aggregates these via cross-attention before proceeding.}
    \label{fig:experts}
\end{figure}

To align experts of different depths (number of layers), we partition each $\mathcal{E}_m$ with $L_m$ transformer layers into $Q$ contiguous stacks (Fig. \ref{fig:experts}): $\mathcal{E}_m = \{S_m^0, S_m^1, \ldots, S_m^{Q-1}\}$ where $|S_m^q| = L_m / Q$.
This enables $Q$ levels of interactions. At the end of each stack, routed experts project their hidden states into a shared latent dimension; the primary expert attends over these projected states and updates its representation before continuing with its forward pass. The backbones remain frozen, while only the router and the projection/attention layers are trained, adding minimal parameter overhead (see Appendix \ref{app:param_overhead}).

\vspace{-23pt}

\subsection{Global Routing}
\label{sec:routing}

\begin{figure}[t]
    \centering
    \includegraphics[width=0.8\linewidth]{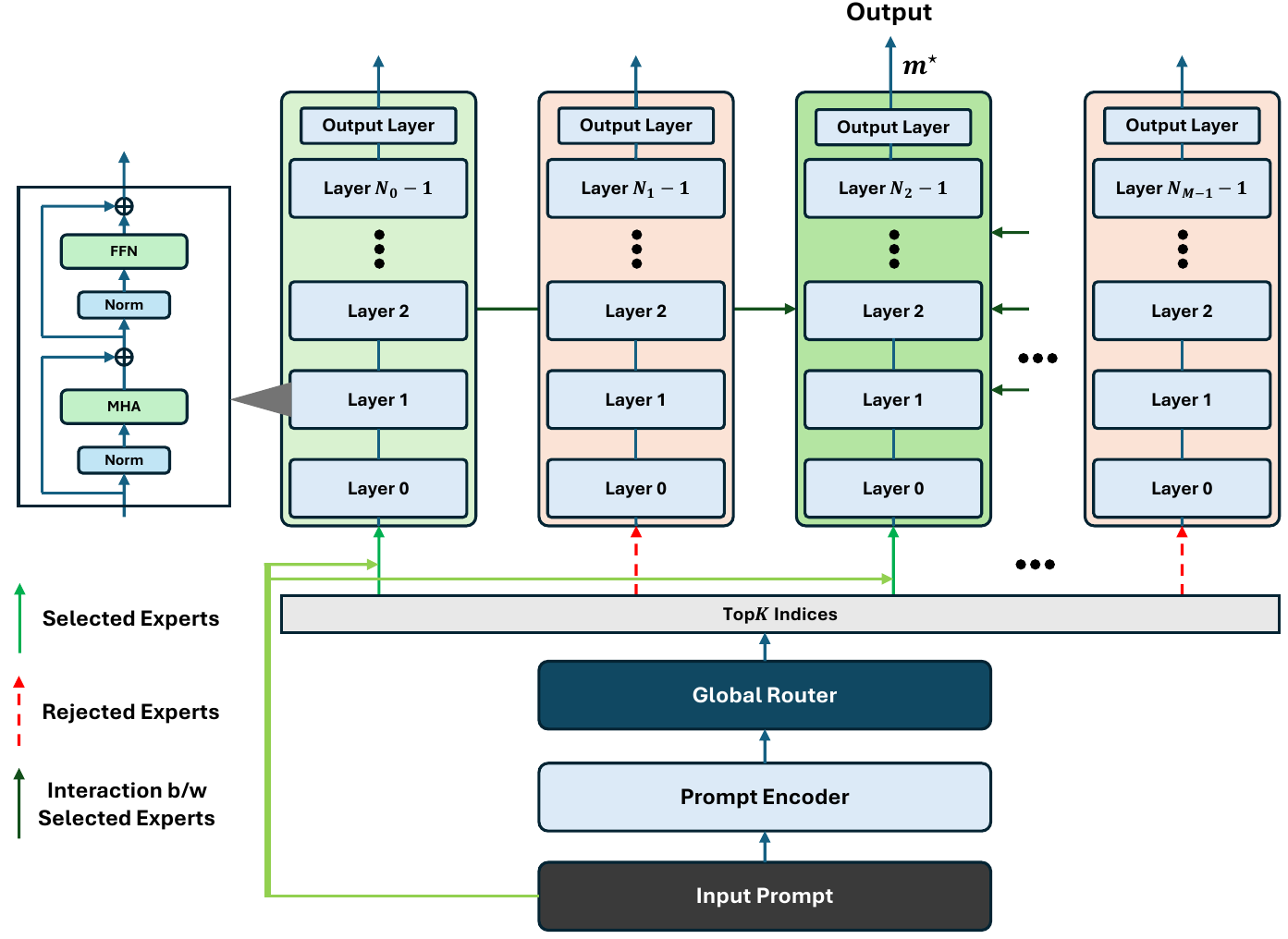}
    \caption{\textbf{Global router.} Each input prompt is encoded, scored against all experts, and the top-$K$ are activated. The highest-scoring expert is designated as the primary decoder responsible for output generation.}
    \label{fig:global}
\end{figure}

The global router (Fig. \ref{fig:global}) selects a sparse subset of experts for each input, ensuring that only the most relevant models participate in collaboration. 
Given an input prompt $\mathbf{x}$, we first compute a fixed-dimensional embedding $\mathbf{z} = \PromptEncoder(\mathbf{x}) \in \mathbb{R}^d$. 
A lightweight MLP router $r_\theta$ maps this embedding to expert relevance scores $\mathbf{s} = r_\theta(\mathbf{z}) \in \mathbb{R}^M$.

The active set of experts is obtained by taking the indices of the top-$K$ scores, $\mathcal{I}_{\text{active}} = \text{TopK}(\mathbf{s}, K)$,
and the highest-scoring expert is designated as the \emph{primary expert}, $m^* = \arg\max_{m \in \mathcal{I}_{\text{active}}} s_m.$
The active experts proceed with their forward pass in parallel, with latent space collaboration via interaction layers at the end of each stack (Fig. \ref{fig:experts}). The final output tokens are generated via the primary expert alone -- integrating information from its active peers, for a given query, via the interaction layers. We utilize a frozen sentence encoder (DeBERTaV3 \cite{debertav3}) to obtain $\mathbf{z}$, while the router $r_\theta$ is trained jointly with the interaction layers (see Sec.~\ref{sec:training}).

\subsection{Interaction Layers}
\label{sec:interaction}

\begin{figure}[t]
    \centering
    \includegraphics[width=0.9\linewidth]{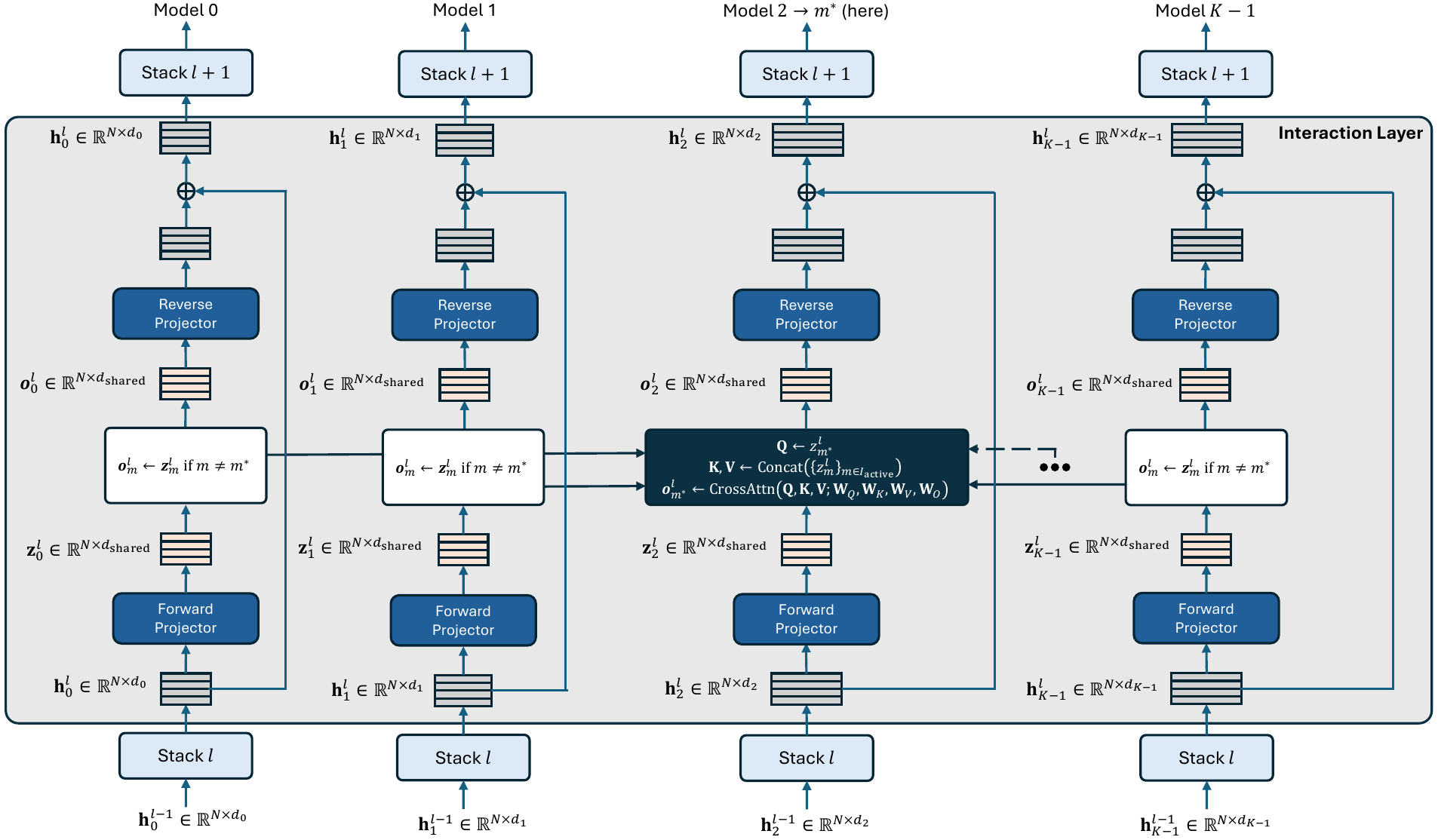}
    \caption{\textbf{Interaction layers.} Active experts project hidden states to a shared space; the primary expert integrates them via cross-attention, while non-primary experts pass through.}
    \label{fig:interaction}
\end{figure}

At the end of each stack $\ell \in \{0,\ldots,Q-1\}$, routed experts collaborate in a shared latent space (Fig.~\ref{fig:interaction}). 
Let $\mathbf{h}_m^\ell \in \mathbb{R}^{N\times d_m}$ be the hidden states of expert $m$ after stack $\ell$. Each active expert applies a $\mathrm{ForwardProjector}$, a linear layer mapping into the shared dimension $d_s$: $\mathbf{z}_m^\ell = \mathrm{ForwardProjector}_m^\ell(\mathbf{h}_m^\ell)$. For the primary expert $m^*$, cross-attention integrates the projected states from all active experts. We set $\text{query}^\ell = \mathbf{z}_{m^*}^\ell$ and $\text{key}^\ell = \text{value}^\ell = \mathrm{concat}\{\mathbf{z}_m^\ell \mid m \in \mathcal{I}_{\text{active}}\}$. After projection via weights $\mathbf{W}_Q^\ell$, $\mathbf{W}_K^\ell$, and $\mathbf{W}_V^\ell$, we get the query ($\mathbf{q}^{\ell}$), key ($\mathbf{k}^{\ell}$), and value ($\mathbf{v}^{\ell}$) matrices, respectively. We compute cross-attention via:
\[
\mathbf{A}^{\ell} = \mathrm{softmax}\!\left(\frac{\mathbf{q}^{\ell}(\mathbf{k}^{\ell})^\top}{\sqrt{d_s}} + \mathbf{M}^\ell\right)\mathbf{v}^{\ell},\qquad
\mathbf{o}_{m^*}^\ell = \mathbf{A}^{\ell}\mathbf{W}_O^\ell.
\]
In practice, we use multi-headed cross-attention \cite{vaswani2017attention} with $H=8$ heads. $\mathbf{M}^\ell \in \mathbb{R}^{N\times (K N)}$ is a block-causal mask ensuring autoregressive visibility across tokens of all active experts (see Appendix \ref{app:inference}). The primary expert hidden states are updated via a $\mathrm{ReverseProjector}$: $\mathbf{h}_{m^*}^\ell \leftarrow \mathbf{h}_{m^*}^\ell + \mathrm{ReverseProjector}_{m^*}^\ell(\mathbf{o}_{m^*}^\ell)$. For each non-primary active expert $m \neq m^*$, no cross-attention is computed. Instead, states pass through the projector pair to keep parameters trainable while avoiding extra compute: $\mathbf{h}_m^\ell \leftarrow \mathbf{h}_m^\ell + \mathrm{ReverseProjector}_m^\ell(\mathbf{z}_m^\ell)$. Thus only the primary expert executes cross-attention, while all experts train their lightweight projectors. This enables frozen backbones to collaborate through a shared latent space at minimal cost. Final token generation is taken from the primary expert, while non-primary experts continue decoding to maintain diverse latent trajectories that feed subsequent interactions in each decoding step. The trainable parameters are the router $\Theta_{\text{router}}$, forward/reverse projectors $\Theta_{\text{proj}}=\{(\mathbf{F}_m^\ell,\mathbf{R}_m^\ell)\}_{m,\ell}$, and per-layer shared attention weights $\Theta_{\text{attn}}=\{(\mathbf{W}_Q^\ell,\mathbf{W}_K^\ell,\mathbf{W}_V^\ell,\mathbf{W}_O^\ell)\}_{\ell}$.

\subsection{Training}
\label{sec:training}

We train only the router and interaction layers while keeping all expert backbones frozen (Sec.~\ref{sec:interaction}). 
The main training objective is the standard autoregressive language modeling loss. 
To stabilize routing, we add two standard regularizers: (i) an entropy term on the pre-selection scores $\mathbf{s}=r_\theta(\mathbf{z})$ to encourage exploration \cite{jang2017gumbel,maddison2017concrete}, and (ii) a load-balancing term that penalizes variance in expert activation frequencies across a batch, as in sparse MoE training \cite{shazeer2017outrageously,fedus2021switch}. 
The full objective combines these components with nonnegative weights. Routing is discrete Top-$K$ in the forward pass. Gradients are obtained using a straight-through estimator: i.i.d.\ Gumbel noise is added to $\mathbf{s}$, a soft Top-$K$ distribution is computed, and backpropagation proceeds through the soft scores while the hard indices are used for expert routing \cite{kool2019stochastic}. Prior work has noted that instability in expert selection can degrade convergence and efficiency \cite{stablemoe,omi2025loadbalancingmixtureexperts}. To address this, we introduce a routing-consistency term that penalizes divergence between outputs obtained from two independently perturbed Top-$K$ selections for the same input. This encourages stability under small routing variations while keeping training aligned with discrete inference. Refer Appendix \ref{app:training} for specific details.

\subsection{Inference}
\label{sec:inference}
Inference in MoT proceeds in two phases. In the \emph{prefill} phase, the input prompt is routed to the top-$K$ experts selected by the global router. Each active expert processes its stacks in parallel, with interaction layers enabling the primary expert to integrate projected hidden states from all active experts via masked cross-attention, while non-primary experts only apply forward/reverse projections without attention. In the \emph{decode} phase, the primary expert generates the next token at each step, while non-primary active experts also produce tokens independently to advance their hidden states. Collaboration continues in the interaction layers, where the primary expert performs cross-attention. Generation ends when the primary expert reaches end of sequence token or the maximum length is reached; if a non-primary expert halts earlier, its outputs are treated as padding and masked in subsequent steps. Interaction layers use lightweight projections and masked cross-attention with standard causal masking, and key–value (KV) pairs can be cached across all experts as in standard LLM inference. This makes MoT compatible with efficient KV caching while supporting collaborative generation. Further details are provided in Appendix \ref{app:inference}.

%% file: files/experiments.tex
\section{Experiments}
\label{sec:exp}

\subsection{Setup}
\paragraph{Models.}
MoT routes among seven 7B to 8B decoder-only LLMs drawn from the Mistral and Llama-3 families, covering general, instruction-tuned, aligned, math-specialized, and Chinese-continued-pretraining variants (details in Appendix ~\ref{app:imp}, \ref{app:train_details}). Our experimental setup closely follows prior work~\cite{routerdc, avengers}.

\paragraph{Datasets.}
Training uses the union of train splits from five in-distribution benchmarks---\textbf{MMLU}~\cite{mmlu}, \textbf{GSM8K}~\cite{gsm8k}, \textbf{CMMLU}~\cite{cmmlu}, \textbf{ARC-C}~\cite{ARC-Challenge}, and \textbf{HumanEval}~\cite{humaneval}---with evaluation on their held-out tests. For GSM8K we use its default split; for the others we follow standard practice (from ~\cite{routerdc, avengers}) with a random 70/30 train/test split. Out-of-distribution (OOD) generalization is assessed only on \textbf{PreAlgebra}~\cite{prealgebra}, \textbf{MBPP}~\cite{mbpp}, and \textbf{C-Eval}~\cite{c-eval}.

\paragraph{Baselines.}
We compare to (i) training-free routers: \emph{Voting} and \emph{CosineClassifier}; (ii) clustering-based routers: \emph{Avengers}~\cite{avengers} and \emph{LoraRetriever}~\cite{loraretriever}; and (iii) supervised routers: \emph{RouterDC}~\cite{routerdc} and \emph{ZOOTER}~\cite{zooter}. All methods use the same expert pool and datasets for evaluation and training. Refer Appendix \ref{app:imp} for implementation details and Appendix \ref{app:train_details} for training details.

\subsection{Results}

\paragraph{In-distribution (ID).}
Table~\ref{tab:id} reports accuracy on five ID benchmarks. \textsc{MoT} attains the best average score (60.53), outperforming the prior SOTA \textsc{Avengers} (60.30) by \(\mathbf{+0.38\%}\) and the strongest single-model baseline (Dolphin-2.9-LLaMA-3-8B; 54.56) by \(\mathbf{+10.95\%}\). Relative to \textsc{RouterDC} (58.54), \textsc{MoT} improves ID average by \(\mathbf{+3.40\%}\). Per-task, \textsc{MoT} achieves the top result on \textsc{MMLU}, \textsc{GSM8K}, \textsc{CMMLU}, and \textsc{HumanEval} (4/5 tasks, wrt routing \& aggregation baselines), with a narrow second place on \textsc{ARC-C}. Despite introducing interaction layers, runtime remains comparable to \textsc{Avengers} (51.8 vs.\ 51.3 minutes on the same hardware).

\begin{table}[!h]
\centering
\caption{In-distribution results (accuracy \%, higher is better). ``Time (m)'' is average end-to-end evaluation minutes under the shared setup.}
\label{tab:id}
\tighttable
\setlength{\tabcolsep}{4.5pt}
{\scriptsize
\renewcommand{\arraystretch}{0.96}%
\begin{tabular*}{\textwidth}{@{\extracolsep{\fill}} @{} l c c c c c A T @{}}
\topaccent
\rowcolor{Hdr}
\textbf{Method} & \textbf{MMLU} & \textbf{GSM8K} & \textbf{CMMLU} & \textbf{ARC-C} & \textbf{HumanEval} & \textbf{Avg} & \textbf{Time (m)} \\
\midrule
\rowcolor{Block}\multicolumn{8}{l}{\textbf{Base models}}\\
Mistral-7B                    & 62.14 & 36.71 & 43.83 & 49.43 & 28.98 & 44.22 & 38.8 \\
MetaMath-Mistral-7B           & 59.86 & 69.63 & 43.83 & 48.30 & 29.80 & 50.28 & 40.5 \\
Zephyr-7B-Beta                & 59.81 & 33.00 & 42.82 & 57.95 & 22.04 & 43.13 & 40.6 \\
Chinese-Mistral-7B            & 57.42 & 41.03 & 49.67 & 43.47 & 21.43 & 42.60 & 40.2 \\
Dolphin-2.6-Mistral-7B        & 60.53 & 52.38 & 43.71 & 52.56 & 45.10 & 50.86 & 42.1 \\
Meta-LLaMA-3-8B               & \best{64.59} & 47.76 & 51.77 & 49.43 & 26.73 & 48.06 & 41.2 \\
Dolphin-2.9-LLaMA-3-8B        & 59.46 & 69.81 & 44.72 & 49.43 & 49.39 & 54.56 & \fast{38.6} \\
\midrule
\rowcolor{Block}\multicolumn{8}{l}{\textbf{Ensembles / routers}}\\
Voting                        & 63.30 & 67.39 & 47.48 & 50.85 & 42.85 & 54.37 & 343.8 \\
CosineClassifier              & 59.72 & 69.03 & 45.47 & 50.57 & 46.33 & 54.22 & 49.7 \\
ZOOTER                        & 60.48 & 66.69 & 45.27 & 53.13 & 44.29 & 53.97 & 47.3 \\
LoraRetriever                 & 63.33 & 66.63 & 51.77 & 57.10 & 40.00 & 55.77 & 46.2 \\
RouterDC                      & 61.07 & 70.32 & 51.77 & 58.52 & 51.02 & 58.54 & 46.8 \\
Avengers                      & 62.81 & 71.56 & 52.58 & \best{60.89} & 53.68 & 60.30 & 51.3 \\
\midrule
\rowcolor{Block}\multicolumn{8}{l}{\textbf{Ours}}\\
\textbf{MoT (ours)}           & 63.14 & \best{72.15} & \best{52.95} & 60.35 & \best{54.08} & \best{60.53} & 51.8 \\
\bottomaccent
\end{tabular*}
}
\vspace{2pt}
\captionsetup{font=small}
\caption*{\footnotesize \textbf{Legend:} \fcolorbox{black!5}{Top}{\strut\ } column best; \fcolorbox{black!5}{Fast}{\strut\ } fastest (lowest) time.}
\end{table}

\vspace{-10pt}

\paragraph{Out-of-distribution (OOD).}
Table~\ref{tab:ood} shows accuracy on \textsc{PreAlgebra}, \textsc{MBPP}, and \textsc{C-EVAL}. \textsc{MoT} achieves the best average (47.92), exceeding \textsc{Avengers} (46.56) by \(\mathbf{+2.92\%}\), \textsc{RouterDC} (45.85) by \(\mathbf{+4.52\%}\), and the strongest single-model baseline (Dolphin-2.9-LLaMA-3-8B; 43.95) by \(\mathbf{+9.03\%}\). \textsc{MoT} is best on all three OOD tasks and maintains runtime on par with \textsc{Avengers} (38.1 vs.\ 37.9 minutes). We attribute the stronger OOD generalization to latent-space collaboration via interaction layers: the primary expert can integrate hidden states from heterogeneous peers, going beyond single/multi-model routing or output-level aggregation.

\begin{table}[!t]
\centering
\caption{Out-of-distribution results (accuracy \%).}
\label{tab:ood}
\tighttable
{\scriptsize
\begin{tabular*}{\textwidth}{@{\extracolsep{\fill}} l c c c A T @{}}
\topaccent
\rowcolor{Hdr}
\textbf{Method} & \textbf{PreAlgebra} & \textbf{MBPP} & \textbf{C-EVAL} & \textbf{Avg} & \textbf{Time (m)} \\
\midrule
\rowcolor{Block}\multicolumn{6}{l}{\textbf{Base models}}\\
Mistral-7B                    & 24.80 & 37.90 & 46.43 & 36.38 & 31.3 \\
MetaMath-Mistral-7B           & 39.15 & 37.74 & 45.17 & 40.69 & 30.6 \\
Zephyr-7B-Beta                & 20.78 & 31.14 & 44.87 & 32.26 & 32.7 \\
Chinese-Mistral-7B            & 18.48 & 29.64 & 48.44 & 32.19 & 32.9 \\
Dolphin-2.6-Mistral-7B        & 29.28 & 44.86 & 45.10 & 39.75 & 28.4 \\
Meta-LLaMA-3-8B               & 27.67 & 43.02 & 52.01 & 40.90 & 27.9 \\
Dolphin-2.9-LLaMA-3-8B        & 39.72 & 47.34 & 44.80 & 43.95 & \fast{27.6} \\
\midrule
\rowcolor{Block}\multicolumn{6}{l}{\textbf{Ensembles / routers}}\\
Voting                        & 39.03 & 41.60 & 48.50 & 43.04 & 205.4 \\
CosineClassifier              & 36.97 & 38.48 & 47.77 & 41.07 & 33.0 \\
ZOOTER                        & 34.44 & 41.10 & 44.95 & 40.16 & 31.6 \\
LoraRetriever                 & 35.36 & 43.12 & 52.01 & 43.50 & 31.2 \\
RouterDC                      & 38.81 & 46.80 & 51.93 & 45.85 & 32.6 \\
Avengers                      & 38.95 & 48.11 & 52.63 & 46.56 & 37.9 \\
\midrule
\rowcolor{Block}\multicolumn{6}{l}{\textbf{Ours}}\\
\textbf{MoT (ours)}           & \best{39.89} & \best{48.56} & \best{55.32} & \best{47.92} & 38.1 \\
\bottomaccent
\end{tabular*}
}
\end{table}

\subsection{Ablations}

\begin{figure*}
    \centering
    \begin{subfigure}[t]{0.32\textwidth}
        \centering
        \includegraphics[width=\linewidth]{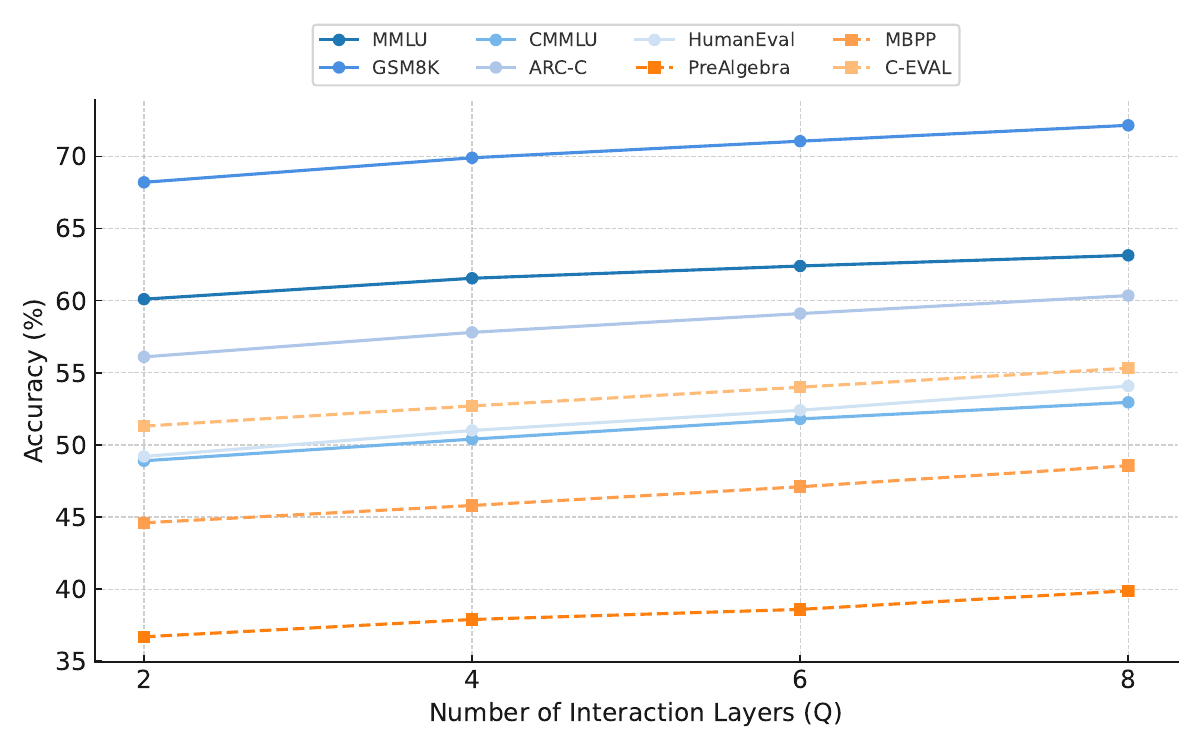}
        \caption{\textbf{Interaction layers ($Q$).} Accuracy vs.\ $Q$ on ID (solid) and OOD (lighter). Gains saturate beyond $Q{=}6$.}
        \label{fig:q-ablation}
    \end{subfigure}
    \hfill
    \begin{subfigure}[t]{0.32\textwidth}
        \centering
        \includegraphics[width=\linewidth]{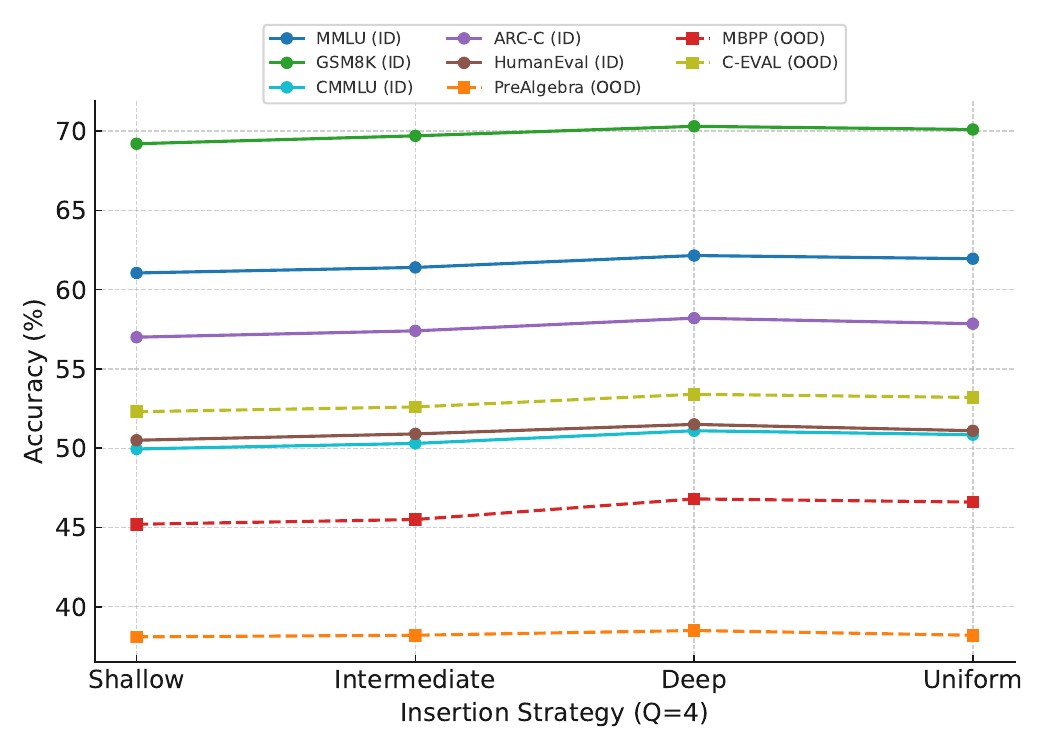}
        \caption{\textbf{Insertion depth.} Accuracy of shallow, intermediate, deep, and uniform placement of $Q{=}4$ layers across ID/OOD tasks.}
        \label{fig:insdepth-detailed}
    \end{subfigure}
    \hfill
    \begin{subfigure}[t]{0.32\textwidth}
        \centering
        \includegraphics[width=\linewidth]{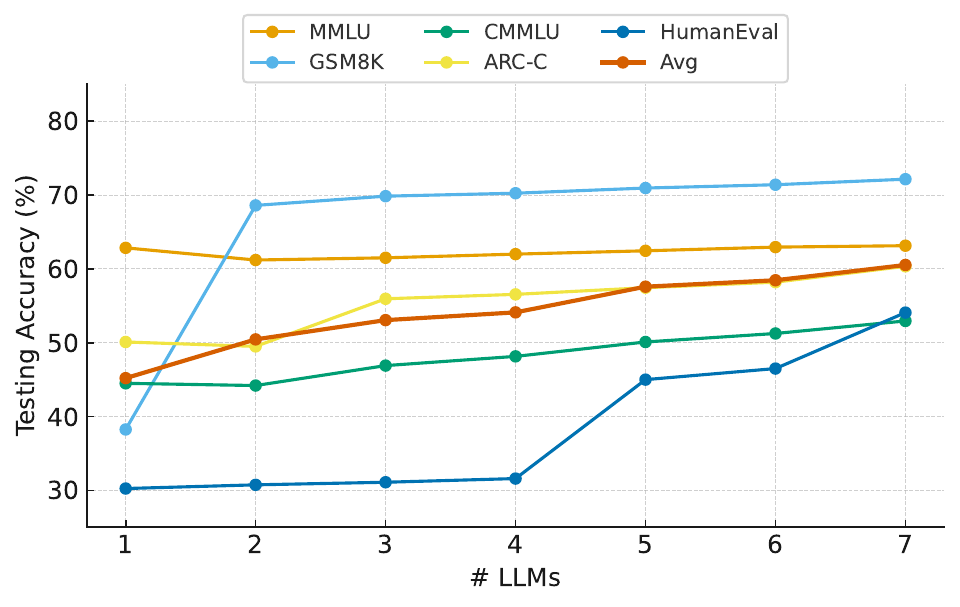}
        \caption{\textbf{Number of experts.} Accuracy on ID tasks as the pool grows from 1 to 7 experts with top-$K{=}3$ routing.}
        \label{fig:num_llms}
    \end{subfigure}
    \caption{\textbf{MoT ablations.} (a) Effect of interaction layers, (b) insertion depth, and (c) number of experts. All variants follow the same training setup as Sec.~\ref{sec:exp}.}
    \label{fig:ablations-grid}
\end{figure*}

\paragraph{Number of interaction layers (\(Q\)).}
We vary \(Q \in \{2,4,6,8\}\) while keeping all other settings fixed (Sec.~\ref{sec:exp}). 
As shown in Fig.~\ref{fig:q-ablation}, accuracy improves consistently as $Q$ increases. 
ID average rises from 56.50 at $Q{=}2$ to 60.53 at $Q{=}8$, while OOD average improves from 44.20 to 47.92 over the same range. 
The trend shows clear benefits from increased collaboration via more interaction layers, but gains diminish beyond $Q{=}6$, reflecting a trade-off between added computation (and parameters) from more interaction layers and incremental accuracy improvements. Refer Appendix \ref{app:num_int} for details.

\paragraph{Insertion depth of interaction layers.}
We next study where to insert $Q{=}4$ interaction layers within each expert: \emph{Shallow} (earliest half of layers), \emph{Intermediate} (middle half), \emph{Deep} (latest half), and a \emph{Uniform} control (even spacing). 
Results in Fig.~\ref{fig:insdepth-detailed} show that deeper placement yields slightly stronger performance (58.65 ID avg., 46.23 OOD avg.), compared to shallow or intermediate placements. 
The uniform control (58.37 ID avg., 46.00 OOD avg.) remains competitive, suggesting that while later-layer collaboration is marginally more effective, uniform placement offers a simple and robust default strategy. Refer Appendix \ref{app:int_inser} for details.

\paragraph{Lightweight MoT.}
To examine the efficiency–performance trade-off, we evaluate a smaller MoT variant with reduced capacity: $Q{=}6$ stacks, shared dimension $d_s{=}1024$, and router hidden size $128$. This configuration substantially reduces parameter count and training cost compared to the default ($Q{=}8$, $d_s{=}2048$, router hidden $768$), while keeping the same number of attention heads ($H{=}8$). As shown in Table~\ref{tab:mot-ablate}, the lightweight MoT maintains strong accuracy on both ID and OOD tasks, achieving only a modest drop relative to the full configuration, while reducing wall-clock inference time by $\sim$32\%. This demonstrates that MoT remains effective even under smaller parameter budgets. For parameter counts, see Appendix \ref{app:param_overhead}. 

\begin{table}[!t]
\centering
\caption{Ablations on MoT size and primary expert cross-attention. Accuracy (\%) on ID and OOD benchmarks. Inference time is wall-clock minutes.}
\label{tab:mot-ablate}
\scalebox{0.95}{%
\resizebox{\linewidth}{!}{%
\begin{tabular}{@{}l c c c c c A c c c A T@{}}
\topaccent
\rowcolor{Hdr}
& \multicolumn{6}{c}{\textbf{In-Distribution (ID)}} & \multicolumn{5}{c}{\textbf{Out-of-Distribution (OOD)}} \\
\cmidrule(lr){2-7}\cmidrule(l){8-12}
\rowcolor{Hdr}
\textbf{Method} & \textbf{MMLU} & \textbf{GSM8K} & \textbf{CMMLU} & \textbf{ARC-C} & \textbf{HumanEval} & \textbf{Avg} & \textbf{PreAlg.} & \textbf{MBPP} & \textbf{C-EVAL} & \textbf{Avg} & \textbf{Time (m)} \\
\midrule
\rowcolor{Block}\multicolumn{12}{l}{\textbf{Ablations}}\\
MoT (base)         & \best{63.14} & \best{72.15} & \best{52.95} & \best{60.35} & \best{54.08} & \best{60.53} & \best{39.89} & \best{48.56} & \best{55.32} & \best{47.92} & 51.8 \\
MoT (small)        & 61.12 & 69.80 & 51.40 & 58.90 & 50.12 & 58.27 & 38.70 & 45.62 & 52.01 & 45.44 & \fast{35.5} \\
MoT w/o CrossAttn  & 62.71 & 69.84 & 51.23 & 58.41 & 50.12 & 58.06 & 38.14 & 47.02 & 53.71 & 46.29 & 49.5 \\
\bottomaccent
\end{tabular}
} 
} 
\end{table}

\paragraph{Removing primary expert cross-attention.}
A natural question is whether MoT’s gains stem merely from the additional parameters in the interaction layers, rather than from latent-space collaboration. To test this, we train a variant where the projection adapters are retained but the cross-attention module in the primary expert is removed, eliminating inter-expert latent exchange while keeping a similar parameter budget. As shown in Table~\ref{tab:mot-ablate}, this variant (\emph{MoT w/o CrossAttn}) yields noticeably lower accuracy on both ID and OOD tasks compared to the full MoT. This confirms that the performance gains are not simply due to added parameters: the cross-attention interaction in the primary expert is critical for enabling effective latent-space collaboration across heterogeneous experts.

\paragraph{Effect of training loss components.}
We ablate the contribution of each training loss in MoT by incrementally adding entropy ($\mathcal{L}_{\text{ent}}$), balance ($\mathcal{L}_{\text{bal}}$), and consistency ($\mathcal{L}_{\text{con}}$) regularizers to the base language modeling (LM) objective. 
As shown in Table~\ref{tab:loss-ablation}, each component improves performance, with entropy regularization stabilizing routing, balance loss promoting even expert utilization, and consistency loss enhancing inter-expert alignment. 
Together, these losses yield a cumulative improvement of \textbf{+7.02\%} over the LM-only baseline. 

\begin{table}[!t]
\centering
\caption{Ablation of training loss components on ID benchmarks. Accuracy (\%) and incremental gains.}
\label{tab:loss-ablation}
\tighttable
{\scriptsize
\begin{tabular*}{\textwidth}{@{\extracolsep{\fill}} l l A c @{}}
\topaccent
\rowcolor{Hdr}
\textbf{Configuration} & \textbf{Loss Components} & \textbf{Accuracy} & \textbf{Gain} \\
\midrule
\rowcolor{Block}\multicolumn{4}{l}{\textbf{Loss ablations}}\\
LM Only               & Language Modeling (LM)                                   & 52.42 & -- \\
LM + Ent              & LM + Entropy Regularization (Entropy)                    & 54.36 & +1.94 \\
LM + Ent + Bal        & LM + Entropy + Load Balancing (Balance)                  & 57.29 & +2.93 \\
LM + Ent + Bal + Con  & LM + Entropy + Balance + Routing Consistency             & \best{59.44} & +2.14 \\
\midrule
\rowcolor{Block}\textbf{Cumulative Gain} & -- & -- & \textbf{+7.02} \\
\bottomaccent
\end{tabular*}
}
\end{table}

\subsection{Analysis}

\paragraph{Robustness to LLM loss during inference.}
In real deployments, model servers may become unavailable due to hardware or network issues, making robustness to expert loss a critical property. We evaluate \textsc{MoT} by removing one expert at inference time while keeping the router and interaction layers unchanged from the original 7-expert training. Table~\ref{tab:dropout} reports ID performance when each expert is dropped in turn. Across all settings, accuracy decreases only marginally (average $-0.57$ compared to full \textsc{MoT}), confirming that the framework gracefully tolerates expert loss. This robustness arises from the fact that multiple experts are routed per query and the primary expert can still integrate information from the remaining active peers through interaction layers. Thus, \textsc{MoT} can maintain high performance even under partial system failures.

\begin{table}[!t]
\centering
\caption{Robustness to expert loss during inference (accuracy \%). Each row drops one expert from the 7-expert pool at test time, while using the same trained router.}
\label{tab:dropout}
\tighttable
{\scriptsize
\setlength{\tabcolsep}{5pt}
\begin{tabular*}{\textwidth}{@{\extracolsep{\fill}} @{} l c c c c c A @{}}
\topaccent
\rowcolor{Hdr}
\textbf{Variant} & \textbf{MMLU} & \textbf{GSM8K} & \textbf{CMMLU} & \textbf{ARC-C} & \textbf{HumanEval} & \textbf{Avg} \\
\midrule
\rowcolor{Block}\multicolumn{7}{l}{\textbf{All experts vs. single drop}}\\
\textbf{All experts}                & \best{63.14} & \best{72.15} & \best{52.95} & \best{60.35} & \best{54.08} & \best{60.53} \\
\midrule
w/o Mistral-7B                      & 62.40 & 71.70 & 52.50 & 59.80 & 53.40 & 59.96 \\
w/o MetaMath-Mistral-7B             & 62.15 & 71.40 & 52.25 & 59.30 & 53.00 & 59.62 \\
w/o Zephyr-7B-Beta                  & 61.95 & 71.10 & 52.10 & 58.95 & 52.70 & 59.36 \\
w/o Chinese-Mistral-7B              & 62.25 & 71.35 & 52.35 & 59.20 & 53.10 & 59.65 \\
w/o Dolphin-2.6-Mistral-7B          & 62.30 & 71.50 & 52.40 & 59.40 & 53.05 & 59.73 \\
w/o Meta-LLaMA-3-8B                 & 61.85 & 71.00 & 51.85 & 58.85 & 52.50 & 59.21 \\
w/o Dolphin-2.9-LLaMA-3-8B          & 61.70 & 70.85 & 51.80 & 58.70 & 52.35 & 59.08 \\
\bottomaccent
\end{tabular*}
}
\end{table}

\paragraph{Effect of the number of experts.}
We study scaling by enabling subsets of the expert pool (1 to 7). 
Fig.~\ref{fig:num_llms} shows accuracy improves steadily with more experts, despite training once on the full pool. 
Top-$K{=}3$ routing remains effective beyond three experts, underscoring MoT’s ability to leverage heterogeneous expertise without retraining. 
Details are in Appendix~\ref{app:num_exp}.

%% file: files/conclusion.tex
\section{Conclusion}
We introduced \emph{Mixture of Thoughts} (\textsc{MoT}), a method that combines global routing with latent-level collaboration to integrate multiple heterogeneous LLMs. By enabling fine-grained hidden-state exchange through lightweight interaction layers while keeping expert backbones frozen, \textsc{MoT} achieves single-pass inference that is both efficient and effective. Experiments on diverse in-distribution and out-of-distribution benchmarks show that \textsc{MoT} consistently outperforms baselines, achieving state-of-the-art performance for multi-LLM systems. 

%% file: files/appendix_new.tex
\section{Implementation Details}
\label{app:imp}
Following previous work, we evaluate MoT using the Language Model Evaluation Harness \cite{eval-harness}. 
Our default MoT configuration employs $Q=8$ stacks, top-$K=3$ routed experts, shared latent dimension $d_s=2048$, router hidden size $768$, and $H=8$ attention heads in the interaction layers. 
The router input encoder is DeBERTaV3-base~\cite{debertav3}, used as a frozen sentence encoder. 
Training uses AdamW with learning rate $5\times 10^{-5}$, batch size $64$, gradient clipping, weight decay $0.01$, and $50$ epochs with $1000$ warmup steps. 
Regularization weights are set to $\lambda_{\text{ent}}=0.01$, $\lambda_{\text{bal}}=0.01$, and $\lambda_{\text{con}}=0.05$. For open-ended generation tasks we sample $M=10$ outputs per query with temperature $0.2$. We train on NVIDIA A100 80 GB GPUs and perform inference on NVIDIA RTX A6000 48 GB GPUs.

\subsection{Models}
We evaluate on seven open-source decoder-only LLMs in the 7B--8B range:
\begin{itemize}
    \item \emph{Mistral family (5):} \textbf{Mistral-7B}~\cite{mistral7b} (general), \textbf{MetaMath-Mistral-7B}~\cite{meta-math} (MetaMathQA-tuned), \textbf{Zephyr-7B-Beta} (DPO-aligned chat), \textbf{Chinese-Mistral-7B} (vocabulary-expanded, continued pretraining on Chinese corpora), and \textbf{dolphin-2.6-mistral-7b} (instruction-tuned, Cognitive Computations).
    \item \emph{Llama-3 family (2):} \textbf{Llama-3-8B}~\cite{llama-3} (general) and \textbf{dolphin-2.9-llama3-8b} (instruction-tuned, Cognitive Computations).
\end{itemize}
This pool mixes general, alignment-focused, domain-specialized, and multilingual experts while keeping scale comparable. 

\subsection{Datasets}
\paragraph{In-distribution.}
(i) \textbf{MMLU}~\cite{mmlu} (57 subjects, broad knowledge);
(ii) \textbf{GSM8K}~\cite{gsm8k} (grade-school math);
(iii) \textbf{CMMLU}~\cite{cmmlu} (67 Chinese subjects);
(iv) \textbf{ARC-Challenge (ARC-C)}~\cite{ARC-Challenge} (multi-choice reasoning); and
(v) \textbf{HumanEval}~\cite{humaneval} (Python code generation).
For GSM8K, we use its default train/test split; for the others, we randomly split 70\%/30\% following standard practice.
The union of all training sets forms the data to train the global router and interaction layers; evaluation is on the held-out test sets.

\paragraph{Out-of-distribution.}
(i) \textbf{PreAlgebra}~\cite{prealgebra} (algebraic problems);
(ii) \textbf{MBPP}~\cite{mbpp} (1{,}000 Python coding tasks); and
(iii) \textbf{C-Eval}~\cite{c-eval} (52 Chinese subjects across four difficulty levels).
These datasets are unseen during training and used only for OOD evaluation.

\subsection{Baselines}
We benchmark against a broad spectrum of routing/aggregation strategies:
\begin{itemize}
    \item \emph{CosineClassifier}: multi-class routing via a cosine-similarity classifier.
    \item \emph{Voting}: query all experts and take the majority output.
    \item \emph{ZOOTER}~\cite{zooter}: supervised router trained with reward-based scoring signals.
    \item \emph{LoraRetriever}~\cite{loraretriever}: routing from query clustering as a proxy for task identity.
    \item \emph{RouterDC}~\cite{routerdc}: supervised router using dual contrastive learning between queries and model embeddings.
    \item \emph{Avengers}~\cite{avengers}: state-of-the-art training-free clustering with ensemble-style sampling and voting.
\end{itemize}
Together, these baselines span simple training-free heuristics and advanced supervised routers under a consistent evaluation protocol.

\section{Training Strategy}
\label{app:training}

We train only lightweight components—global router $r_\theta$, per-expert/per-stack forward/reverse projectors, and shared cross-attention weights—while keeping all expert backbones frozen. 

\medskip
\noindent \textbf{Primary objective.}
For input sequence $\mathbf{x}$ and target sequence $\mathbf{y}=(y_0,\ldots,y_{T-1})$, the primary expert $m^*$ predicts next-token distributions $p_{m^*}(y_t\mid\mathbf{x},\mathbf{y}_{<t})$. We minimize the autoregressive negative likelihood loss (NLL):
\[
\mathcal{L}_{\mathrm{LM}} \;=\; -\sum_{t=0}^{T-1}\log p_{m^*}\!\left(y_t \mid \mathbf{x}, \mathbf{y}_{<t}\right).
\]

\medskip
\noindent \textbf{Router exploration (entropy).}
Let $\mathbf{z}=\PromptEncoder(\mathbf{x})$ and $\mathbf{s}=r_\theta(\mathbf{z})\in\mathbb{R}^M$ be router scores. To discourage collapse onto a few experts, we maximize the entropy of distribution $\pi_\tau=\mathrm{softmax}(\mathbf{s}/\tau)$, implemented as a loss:
\[
\mathcal{L}_{\mathrm{ent}} \;=\; -\,\mathbb{E}_{\mathbf{x}}\!\left[\,\mathcal{H}\!\left(\pi_\tau\right)\right].
\]
We use a straight-through (ST) relaxation: \emph{hard} Top-$K$ indices are used in the forward pass; gradients flow through the \emph{soft} $\pi_\tau$ in the backward pass. A standard choice is Gumbel-perturbed scores $\tilde{\mathbf{s}}=\mathbf{s}+\mathbf{g}$ (i.i.d.\ $\mathbf{g}$ Gumbel) with temperature $\tau$ \cite{jang2017gumbel, maddison2017concrete}, yielding stable SoftTopK relaxations while preserving discrete routing at inference.

\medskip
\noindent \textbf{Load balancing.}
Within a batch of size $B$, let $f_m=\tfrac{1}{B}\sum_{i=0}^{B-1} \mathbf{1}[\,m\in \mathcal{I}^{(i)}_{\mathrm{active}}\,]$ denote the activation frequency of expert $m$. We penalize the squared coefficient of variation to encourage uniform utilization \cite{shazeer2017outrageously, fedus2021switch}:
\[
\mathcal{L}_{\mathrm{bal}} \;=\; \Big(\tfrac{\mathrm{std}(\mathbf{f})}{\mathrm{mean}(\mathbf{f})}\Big)^2,
\qquad \mathbf{f}=(f_0,\ldots,f_{M-1}).
\]

\medskip
\noindent \textbf{Routing-consistency (stability under small routing changes).}
Discrete Top-$K$ selections can vary due to stochastic perturbations; we regularize for stability by sampling two \emph{independent} relaxed selections for the same input. Concretely, draw $\mathbf{g}_0,\mathbf{g}_1\sim\mathrm{Gumbel}^M$ and form $\tilde{\mathbf{s}}_j=(\mathbf{s}+\mathbf{g}_j)/\tau$ for $j\in\{0,1\}$. We use \emph{hard} Top-$K$ from $\tilde{\mathbf{s}}_0$ for the forward pass (primary distribution $p^{(0)}_t$), and in parallel we compute a distribution $p^{(1)}_t$ using experts selected via $\tilde{\mathbf{s}}_1$. We minimize token-wise KL divergence between the two distributions over the target tokens:
\[
\mathcal{L}_{\mathrm{con}} \;=\; \frac{1}{2T}\sum_{t=0}^{T-1}\Big(
\mathrm{KL}\!\big(p^{(0)}_t \,\|\, p^{(1)}_t\big)\;+\;\mathrm{KL}\!\big(p^{(1)}_t \,\|\, p^{(0)}_t\big)
\Big).
\]

\medskip
\noindent \textbf{Total objective.}
The full loss combines the above terms:
\[
\mathcal{L}_{\mathrm{total}}
\;=\;
\mathcal{L}_{\mathrm{LM}}
\;+\;\lambda_{\mathrm{ent}}\,\mathcal{L}_{\mathrm{ent}}
\;+\;\lambda_{\mathrm{bal}}\,\mathcal{L}_{\mathrm{bal}}
\;+\;\lambda_{\mathrm{con}}\,\mathcal{L}_{\mathrm{con}}.
\]
We use AdamW with weight decay, gradient clipping, and warmup. Weights $(\lambda_{\mathrm{ent}},\lambda_{\mathrm{bal}},\lambda_{\mathrm{con}})$ are fixed; the defaults used in our experiments are $\lambda_{\mathrm{ent}}=0.01$, $\lambda_{\mathrm{bal}}=0.01$, $\lambda_{\mathrm{con}}=0.05$.

\begin{algorithm}
\caption{MoT Training (single batch step)}
\label{alg:mot-train}
\begin{algorithmic}[1]
\STATE \textbf{Inputs:} batch $(\mathbf{x},\mathbf{y})$, router $r_\theta$, projectors $\Theta_{\text{proj}}$, cross-attn $\Theta_{\text{attn}}$, frozen experts $\{\mathcal{E}_m\}_{m=0}^{M-1}$
\STATE Encode prompt: $\mathbf{z} \leftarrow \PromptEncoder(\mathbf{x})$; \quad $\mathbf{s} \leftarrow r_\theta(\mathbf{z})$
\STATE Sample $\mathbf{g}_0 \!\sim\! \mathrm{Gumbel}^M$, form $\tilde{\mathbf{s}}_0=(\mathbf{s}+\mathbf{g}_0)/\tau$, select $\mathcal{I}_{\text{act}}=\text{TopK}(\tilde{\mathbf{s}}_0)$
\STATE Run active experts with interaction layers; primary $m^*$ produces distributions $\{p^{(0)}_t\}_{t=0}^{T-1}$; compute $\mathcal{L}_{\mathrm{LM}}$
\STATE Compute router entropy loss $\mathcal{L}_{\mathrm{ent}}=-\mathcal{H}(\mathrm{softmax}(\mathbf{s}/\tau))$
\STATE Compute load balance loss $\mathcal{L}_{\mathrm{bal}}=\big(\mathrm{std}(\mathbf{f})/\mathrm{mean}(\mathbf{f})\big)^2$ with $\mathbf{f}=(f_0,\ldots,f_{M-1})$
\STATE Sample $\mathbf{g}_1 \!\sim\! \mathrm{Gumbel}^M$, form $\tilde{\mathbf{s}}_1=(\mathbf{s}+\mathbf{g}_1)/\tau$, select $\mathcal{I}'_{\text{act}}=\text{TopK}(\tilde{\mathbf{s}}_1)$
\STATE Run secondary active experts $\mathcal{I}'_{\text{act}}$ with interaction layers; obtain $\{p^{(1)}_t\}_{t=0}^{T-1}$; compute $\mathcal{L}_{\mathrm{con}}$
\STATE Combine: $\mathcal{L}_{\mathrm{total}}=\mathcal{L}_{\mathrm{LM}}+\lambda_{\mathrm{ent}}\mathcal{L}_{\mathrm{ent}}+\lambda_{\mathrm{bal}}\mathcal{L}_{\mathrm{bal}}+\lambda_{\mathrm{con}}\mathcal{L}_{\mathrm{con}}$
\STATE Update trainable parameters $\{r_\theta,\Theta_{\text{proj}},\Theta_{\text{attn}}\}$ (experts frozen) using AdamW
\end{algorithmic}
\end{algorithm}

\section{Inference}
\label{app:inference}

\noindent
MoT inference proceeds in two phases: \emph{prefill} and \emph{decode}. The router selects a sparse set of experts; all selected experts advance their own token streams to maintain latent trajectories, while the \emph{primary} expert produces the final output sequence. Standard causal masking is used in the cross-attention computed within the interaction layers.

\medskip
\noindent \textbf{Prefill.}
Given input $\mathbf{x}$, compute $\mathbf{z}=\PromptEncoder(\mathbf{x})$, scores $\mathbf{s}=r_\theta(\mathbf{z})$, select $\mathcal{I}_{\text{act}}=\text{TopK}(\mathbf{s})$, and choose the primary $m^*=\arg\max_{m\in\mathcal{I}_{\text{act}}} s_m$. Run a forward pass of all active experts on $\mathbf{x}$ with interaction layers enabled. Cache standard backbone KV pairs and the interaction-layer KV projections at every stack for each active expert. Each active expert emits its next token; the primary token is appended to the output sequence.

\medskip
\noindent \textbf{Decode.}
At step $t\!\ge\!0$, each active expert $m$ receives its own previous token (the primary expert receives the last output token). Using KV caches, the tokens traverse through the stacks. At each interaction layer, every active expert updates its KV cache; the primary expert's query token attends over the concatenated KV from all active experts to update its hidden state. New tokens are produced for all active experts; only the primary token is appended to the output. Decoding continues until the primary emits \texttt{<EOS>} or the maximum sequence length is reached. If a non-primary expert emits \texttt{<EOS>} earlier, its subsequent tokens are treated as padding and are masked in the interaction layers.

\begin{figure}
    \centering
    \includegraphics[width=\linewidth]{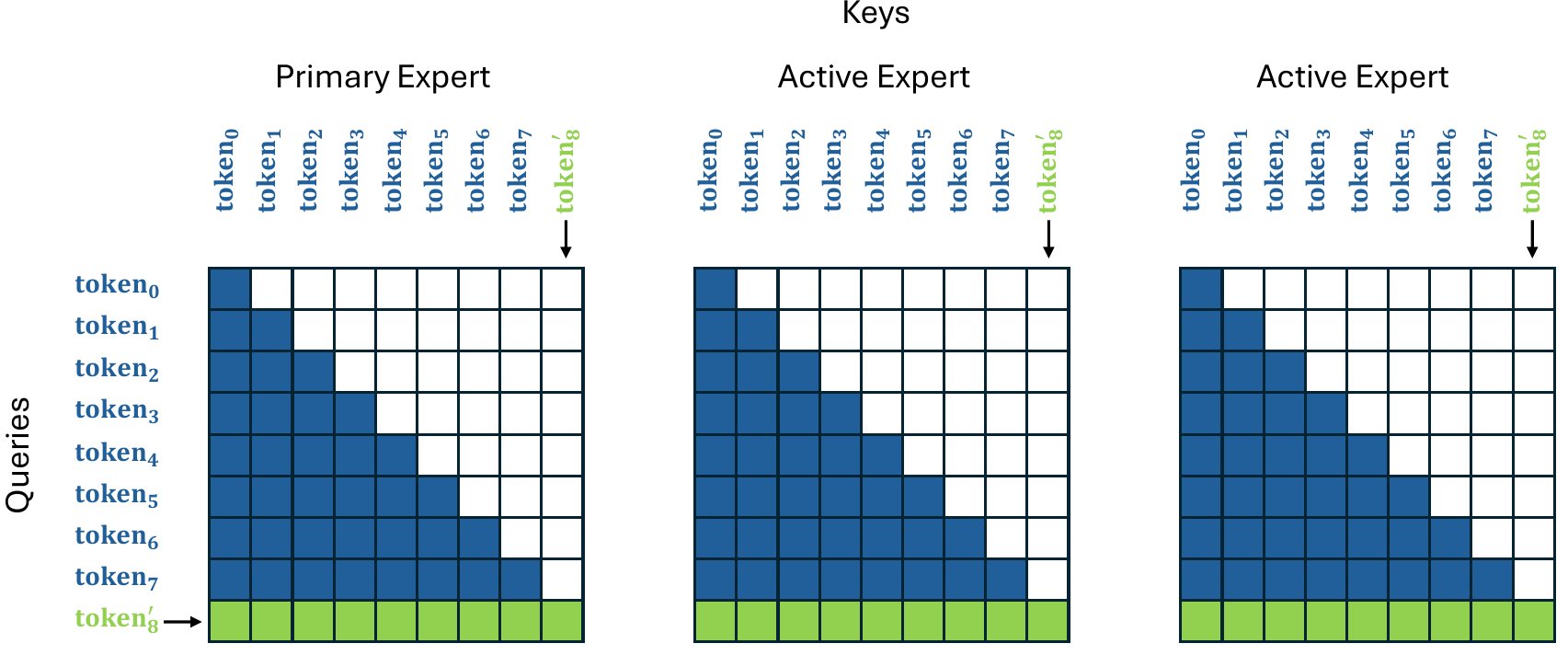}
    \caption{Attention mask used for the cross-attention computation in the interaction layers. The cross-attention is computed between queries from the primary expert and keys/values from the active experts (including the primary expert). The generated token from each active expert after prefill (shown in green) add to the KV cache for the interaction layer.}
    \label{fig:masked_attn_viz}
\end{figure}

\begin{algorithm}
\caption{MoT Inference (single sequence)}
\label{alg:mot-infer}
\begin{algorithmic}[1]
\STATE \textbf{Input:} prompt $\mathbf{x}$, experts $\{\mathcal{E}_m\}_{m=0}^{M-1}$ (frozen), router $r_\theta$, projectors $\Theta_{\text{proj}}$, cross-attn $\Theta_{\text{attn}}$, max length $T_{\max}$
\STATE \textbf{Route:} $\mathbf{z}\!\leftarrow\!\PromptEncoder(\mathbf{x})$;\; $\mathbf{s}\!\leftarrow\! r_\theta(\mathbf{z})$;\; $\mathcal{I}_{\text{act}}\!\leftarrow\!\text{TopK}(\mathbf{s})$;\; $m^*\!\leftarrow\!\arg\max_{m\in\mathcal{I}_{\text{act}}} s_m$
\STATE \textbf{Prefill:} run all $m\!\in\!\mathcal{I}_{\text{act}}$ on $\mathbf{x}$ with interaction layers; cache backbone KV and interaction-layer KV at each stack; obtain next-token logits for each active expert; append the primary token to output $\mathbf{y}$
\STATE Initialize step $t\!\leftarrow\!0$;\; mark all $m\!\in\!\mathcal{I}_{\text{act}}$ as \texttt{alive}
\WHILE{$m^*$ is \texttt{alive} \AND $|\mathbf{y}| < T_{\max}$}
  \FOR{each $m \in \mathcal{I}_{\text{act}}$}
    \IF{$m$ is \texttt{alive}}
      \STATE feed previous token of expert $m$ through its stacks using KV caches
      \STATE at each interaction layer: update KV cache for expert $m$; if $m=m^*$, also form the query token using the current primary token and attend over concatenated KV from all active experts (padding masked)
      \STATE produce next token for expert $m$ (for $m^*$ this is the output token)
      \IF{token is \texttt{<EOS>} and $m\neq m^*$} \STATE mark $m$ as \texttt{dead} \ENDIF
    \ELSE
      \STATE treat token as padding for $m$ (no compute; masked in interaction)
    \ENDIF
  \ENDFOR
  \STATE append the primary token to $\mathbf{y}$;\; if \texttt{<EOS>} for $m^*$ then stop
  \STATE $t \leftarrow t+1$
\ENDWHILE
\STATE \textbf{Return} $\mathbf{y}$
\end{algorithmic}
\end{algorithm}

\clearpage
\section{Results for number of interaction layers}
\label{app:num_int}
\begin{table}[h]
\centering
\caption{Effect of the number of stacks $Q$ (uniformly spaced interaction layers) on in-distribution (ID) and out-of-distribution (OOD) accuracy (\%, higher is better). All runs keep the expert pool, router, and decoding settings fixed; only $Q$ varies.}
\label{tab:stacks-ablation}
\tighttable
\resizebox{\linewidth}{!}{%
\begin{tabular}{@{}l *{5}{A} | A | *{3}{A} | A @{}}
\topaccent
& \multicolumn{6}{c|}{\textbf{In-Distribution (ID)}} & \multicolumn{4}{c}{\textbf{Out-of-Distribution (OOD)}} \\
\cmidrule(lr){2-7}\cmidrule(l){8-11}
\rowcolor{Hdr}
\textbf{$Q$} & \textbf{MMLU} & \textbf{GSM8K} & \textbf{CMMLU} & \textbf{ARC-C} & \textbf{HumanEval} & \textbf{Avg} & \textbf{PreAlg.} & \textbf{MBPP} & \textbf{C-EVAL} & \textbf{Avg} \\
\midrule
2 & 60.10 & 68.20 & 48.90 & 56.10 & 49.20 & 56.50 & 36.70 & 44.60 & 51.30 & 44.20 \\
4 & 61.55 & 69.90 & 50.40 & 57.80 & 51.00 & 58.33 & 37.90 & 45.80 & 52.70 & 45.47 \\
6 & 62.40 & 71.05 & 51.80 & 59.10 & 52.40 & 59.35 & 38.60 & 47.10 & 54.00 & 46.57 \\
8 & \best{63.14} & \best{72.15} & \best{52.95} & \best{60.35} & \best{54.08} & \best{60.53} & \best{39.89} & \best{48.56} & \best{55.32} & \best{47.92} \\
\bottomaccent
\end{tabular}}
\end{table}

\section{Results for interaction layer insertion depth}
\label{app:int_inser}
\begin{table}[h]
\centering
\caption{Effect of interaction-layer insertion depth with $Q{=}4$ stacks. \emph{Shallow:} earliest half of layers; \emph{Intermediate:} middle half; \emph{Deep:} latest half; \emph{Uniform:} evenly spaced control. Accuracy in \% (higher is better).}
\label{tab:depth-ablation}
\tighttable
\resizebox{\linewidth}{!}{%
\begin{tabular}{@{}l *{5}{A} | A | *{3}{A} | A @{}}
\topaccent
& \multicolumn{6}{c|}{\textbf{In-Distribution (ID)}} & \multicolumn{4}{c}{\textbf{Out-of-Distribution (OOD)}} \\
\cmidrule(lr){2-7}\cmidrule(l){8-11}
\rowcolor{Hdr}
\textbf{Insertion} & \textbf{MMLU} & \textbf{GSM8K} & \textbf{CMMLU} & \textbf{ARC-C} & \textbf{HumanEval} & \textbf{Avg} & \textbf{PreAlg.} & \textbf{MBPP} & \textbf{C-EVAL} & \textbf{Avg} \\
\midrule
Shallow        & 61.05 & 69.20 & 49.95 & 57.00 & 50.50 & 57.74 & 38.10 & 45.20 & 52.30 & 45.20 \\
Intermediate   & 61.40 & 69.70 & 50.30 & 57.40 & 50.90 & 58.14 & 38.20 & 45.50 & 52.60 & 45.43 \\
Deep           & \best{62.15} & \best{70.30} & \best{51.10} & \best{58.20} & \best{51.50} & \best{58.65} & \best{38.50} & \best{46.80} & \best{53.40} & \best{46.23} \\
\midrule
\rowcolor{Block}
Uniform (control) & 61.95 & 70.10 & 50.85 & 57.85 & 51.10 & 58.37 & 38.20 & 46.60 & 53.20 & 46.00 \\
\bottomaccent
\end{tabular}}
\end{table}

\section{Results for number of experts}
\label{app:num_exp}
\begin{table}[h]
\centering
\caption{Performance of sequential MoT-style model compositions. Reported metrics are accuracy (\%) on standard benchmarks. The final column is the average across all tasks.}
\label{tab:composition}
\tighttable
\resizebox{\linewidth}{!}{%
\begin{tabular}{@{}l c *{6}{A} @{}}
\topaccent
\rowcolor{Hdr}
\textbf{Model Composition} & \textbf{\#LLMs} & \textbf{MMLU} & \textbf{GSM8K} & \textbf{CMMLU} & \textbf{ARC-C} & \textbf{HumanEval} & \textbf{Avg} \\
\midrule
Mistral-7B                          & 1 & 62.85 & 38.25 & 44.50 & 50.10 & 30.25 & 45.20 \\
+ MetaMath-Mistral-7B               & 2 & 61.20 & 68.60 & 44.20 & 49.50 & 30.75 & 50.45 \\
+ Zephyr-7B-Beta                    & 3 & 61.50 & 69.85 & 46.90 & 55.95 & 31.10 & 53.06 \\
+ Chinese-Mistral-7B                & 4 & 62.00 & 70.25 & 48.15 & 56.55 & 31.60 & 54.11 \\
+ Dolphin-2.6-Mistral-7B            & 5 & 62.45 & 70.95 & 50.10 & 57.45 & 45.00 & 57.59 \\
+ Meta-Llama-3-8B                   & 6 & 62.95 & 71.40 & 51.25 & 58.20 & 46.50 & 58.46 \\
\rowcolor{Block}
+ Dolphin-2.9-Llama3-8B             & 7 & \best{63.14} & \best{72.15} & \best{52.95} & \best{60.35} & \best{54.08} & \best{60.53} \\
\bottomaccent
\end{tabular}}
\end{table}

\section{Parameter Overhead}
\label{app:param_overhead}

For $M$ experts with hidden sizes $\{d_0,\ldots,d_{M-1}\}$, $Q$ stacks (and thus $Q$ interaction layers), shared latent dimension $d_s$, and a frozen prompt encoder that outputs a $d_z$-dimensional vector. The router has two layers with hidden width $h_r$ producing $M$ scores. We count the additional (trainable) parameters due to MoT (omitting biases and layer normalization parameters for simplicity).

\paragraph{Router.}
A 2-layer MLP $d_z \rightarrow h_r \rightarrow M$:
\[
\boxed{\#\theta_{\text{router}} \;=\; d_z h_r \;+\; h_r M~}
\]

\paragraph{Per-expert forward/reverse projectors (per stack).}
At each interaction layer $q\!\in\!\{0,\ldots,Q\!-\!1\}$ and for each expert $m$, we use a forward projector $d_m\!\times\! d_s$ and a reverse projector $d_s\!\times\! d_m$:
\[
\#\theta_{\text{proj per layer}} \;=\; \sum_{m=0}^{M-1}\!\big(d_m d_s + d_s d_m\big) \;=\; 2 d_s \sum_{m=0}^{M-1}\! d_m,
\]
\[
\boxed{~\#\theta_{\text{proj total}} \;=\; Q \cdot \Big( 2 d_s \sum_{m=0}^{M-1}\! d_m \Big)~}
\]

\paragraph{Shared cross-attention per interaction layer.}
We parameterize multi-head attention with four dense projections in the shared space: $W_Q, W_K, W_V, W_O \in \mathbb{R}^{d_s\times d_s}$:
\[
\#\theta_{\text{attn per layer}} \;=\; 4 d_s^2,
\qquad
\boxed{~\#\theta_{\text{attn total}} \;=\; Q \cdot 4 d_s^2~}
\]

\paragraph{Total overhead.}
Summing the three components:
\[
\boxed{~
\#\theta_{\text{total}}
\;=\;
\underbrace{d_z h_r + h_r M}_{\text{router}}
\;+\;
\underbrace{Q \cdot \Big( 2 d_s \sum_{m=0}^{M-1} d_m \Big)}_{\text{proj. (all experts, all stacks)}}
\;+\;
\underbrace{Q \cdot 4 d_s^2}_{\text{cross-attn (all stacks)}}
~}
\]

The added parameters scale (i) linearly with the number of experts via the projector terms, weighted by their hidden sizes $d_m$, (ii) linearly with the number of stacks $Q$, and (iii) quadratically with the shared latent size $d_s$ through the cross-attention block.

\begin{table}[h]
\centering
\caption{Parameter counts of individual models used in composition. Values are reported in billions (B).}
\label{tab:params}
\tighttable
\begingroup
\setlength{\tabcolsep}{4pt}       
\renewcommand{\arraystretch}{1.03}
\small                             
\begin{tabular}{@{}l r@{}}
\topaccent
\rowcolor{Hdr}
\textbf{Model} & \textbf{Parameters (B)} \\
\midrule
Mistral-7B-v0.1         & 6.30 \\
MetaMath-Mistral-7B     & 6.30 \\
Zephyr-7B-Beta          & 6.30 \\
Dolphin-2.6-Mistral-7B  & 6.30 \\
Chinese-Mistral-7B-v0.1 & 6.56 \\
Meta-Llama-3-8B         & 6.99 \\
Dolphin-2.9-Llama3-8B   & 6.99 \\
\midrule
\rowcolor{Block}\textbf{Total} & \textbf{45.75} \\
\bottomaccent
\end{tabular}
\endgroup
\end{table}

\begin{table}[h]
\centering
\caption{Additional trainable parameters for MoT configurations (in \textbf{billions}, B). Overhead is computed with respect to the frozen base pool of experts (45.75B; see Table~\ref{tab:params}).}
\label{tab:mot_overhead_configs}
\tighttable
\begingroup
\setlength{\tabcolsep}{4pt}
\renewcommand{\arraystretch}{1.03}
\small
\begin{tabular}{@{}l r r r r@{}}
\topaccent
\rowcolor{Hdr}
\textbf{Config} & \textbf{Interaction (B)} & \textbf{Router (B)} & \textbf{Total Add (B)} & \textbf{Overhead \%} \\
\midrule
MoT ($Q{=}2$)   & 0.386 & 0.001     & 0.387 & 0.85\% \\
MoT ($Q{=}4$)   & 0.772 & 0.001     & 0.773 & 1.69\% \\
MoT ($Q{=}6$)   & 1.158 & 0.001     & 1.159 & 2.53\% \\
MoT ($Q{=}8$)   & 1.544 & 0.001     & 1.545 & 3.38\% \\
MoT (small)     & 0.466 & $<\!0.001$ & 0.466 & 1.02\% \\
\bottomaccent
\end{tabular}
\endgroup
\end{table}

Table~\ref{tab:params} lists the frozen backbone models used in our system (totaling 45.75B parameters). 
On top of this pool, MoT introduces only lightweight trainable components. 
Table~\ref{tab:mot_overhead_configs} reports the additional parameters due to the router and interaction layers across different stack depths. MoT with $Q=8$ is the base configuration used with shared dimension $d_s=2048$, router hidden size $h_r = 768$, Top-$K = 3$. 
We also experiment with smaller variants $Q=2,4,6$, and an MoT (small) configuration that uses reduced dimensions $Q = 6$, $d_s = 1024$, $h_r = 128$, Top-$K = 3$. MoT adds only 0.8\% to 3.4\% additional parameters in comparison to the size of the frozen expert pool.

\section{Training Details}
\label{app:train_details}

\noindent
This section summarizes the training settings used in our runs. All expert backbones are frozen; only the router, per-expert forward/reverse projectors, and shared cross-attention weights are trained. Details are covered in Tables \ref{tab:train_mot_config} through \ref{tab:eval_settings}.

\paragraph{Expert pool (frozen).}
\begin{itemize}[leftmargin=1.25em,itemsep=2pt,topsep=2pt]
\item \texttt{mistralai/Mistral-7B-v0.1}
\item \texttt{meta-math/MetaMath-Mistral-7B}
\item \texttt{HuggingFaceH4/zephyr-7b-beta}
\item \texttt{cognitivecomputations/dolphin-2.6-mistral-7b}
\item \texttt{itpossible/Chinese-Mistral-7B-v0.1}
\item \texttt{meta-llama/Meta-Llama-3-8B}
\item \texttt{cognitivecomputations/dolphin-2.9-llama3-8b}
\end{itemize}

\begin{table}[h]
\centering
\caption{MoT base configuration used in training. The base configuration matches our main experiments.}
\label{tab:train_mot_config}
\tighttable
\setlength{\tabcolsep}{8pt}
\begin{tabular}{@{} l l AT @{}}
\topaccent
\rowcolor{Hdr}
\textbf{Setting} & \textbf{Value} \\
\midrule
Stacks ($Q$) & 8 \\
Top-$K$ & 3 \\
Shared latent dim ($d_s$) & 2048 \\
Interaction heads & 8 \\
Router hidden size ($h_r$) & 768 \\
Router temperature ($\tau$) & 1.0 \\
Dropout (interaction layers) & 0.1 \\
Prompt encoder (frozen) & \texttt{microsoft/deberta-v3-base} \\
\bottomaccent
\end{tabular}
\end{table}

\begin{table}[h]
\centering
\caption{Optimization hyperparameters.}
\label{tab:train_optim}
\tighttable
\setlength{\tabcolsep}{6pt}
\begin{tabular}{@{} l l l l AT @{}}
\topaccent
\rowcolor{Hdr}
\textbf{Hyperparameter} & \textbf{Value} & \textbf{Hyperparameter} & \textbf{Value} \\
\midrule
Epochs & 50 & Batch size & 64 \\
Grad. accumulation & 1 & Learning rate & $5\times 10^{-5}$ \\
Weight decay & 0.01 & Warmup steps & 1000 \\
Grad clip (L2) & 1.0 & Max length (tokens) & 512 \\
Precision & FP16 & Seed & 42 \\
\bottomaccent
\end{tabular}
\end{table}

\begin{table}[h]
\centering
\caption{Loss weights used during training.}
\label{tab:train_losses}
\tighttable
\setlength{\tabcolsep}{6pt}
\begin{tabular}{@{} l c AT @{}}
\topaccent
\rowcolor{Hdr}\textbf{Component} & \textbf{Weight} \\
\midrule
Entropy regularization ($\lambda_{\mathrm{ent}}$) & 0.01 \\
Load balancing ($\lambda_{\mathrm{bal}}$) & 0.01 \\
Routing consistency ($\lambda_{\mathrm{con}}$) & 0.05 \\
\bottomaccent
\end{tabular}
\end{table}

\begin{table}[h]
\centering
\caption{Decoding/evaluation settings.}
\label{tab:eval_settings}
\tighttable
\setlength{\tabcolsep}{5pt}
\begin{tabular}{@{} l l l l AT @{}}
\topaccent
\rowcolor{Hdr}\textbf{Parameter} & \textbf{Value} & \textbf{Parameter} & \textbf{Value} \\
\midrule
Beam size / samples & 10 & Temperature & 0.2 \\
Top-$p$ & 0.9 & Max new tokens & 256 \\
Return sequences & 1 &  &  \\
\bottomaccent
\end{tabular}
\end{table}

\FloatBarrier

\section{Additional Experiments}
\label{sec:standard-training}

\textbf{Motivation.} A potential concern with multi-benchmark evaluations (Section~\ref{sec:exp}) is overlap between training and evaluation data. Such leakage could artificially inflate performance on both ID and OOD tasks. To address this, we conduct additional experiments following the same setup as before, but with separation between training and evaluation datasets. This ensures that our results fairly demonstrate the advantages of MoT over prior approaches.

\paragraph{Training datasets.} 
For training we use the following datasets: commonsense and multi-choice QA (35\% of steps; HellaSwag, CommonsenseQA, ARC-Challenge), math and quantitative reasoning (25\%; GSM8K, SVAMP), code generation (25\%; MBPP and a 10k subset of APPS), and reading comprehension (15\%; SQuAD v2).

\paragraph{Sampling and batching.}
Percentages above denote the share of \emph{optimizer updates} allocated to each domain: at each update we sample a domain with probabilities $(0.35, 0.25, 0.25, 0.15)$ and then draw the batch from a dataset within that domain using temperature-based sampling (temperature $=2.0$). We use a sequence length of $2048$ and a batch size of 64.


\paragraph{Evaluation.}
We train and evaluate using the same seven 7B to 8B experts. Training setting is identical to that in Sec.~\ref{sec:standard-training}. We report accuracy for benchmarks that are \emph{not} used for training.

\begin{table}[h]
\centering
\caption{\textbf{Additional experiments (standard training configuration).} Accuracy (\%, higher is better) on evaluation benchmarks not used for MoT training. 
For baselines, ``ID'' refers to datasets included in their training; ``OOD'' refers to held-out datasets. 
For MoT, \emph{all} evaluation datasets are OOD since none were used for training.}
\label{tab:addl_standard_results}
\tighttable
\resizebox{\linewidth}{!}{%
\begin{tabular}{@{}l *{3}{A} | A | *{2}{A} | A @{}}
\topaccent
& \multicolumn{4}{c|}{\textbf{Baselines: In-Distribution (ID)}} & \multicolumn{3}{c}{\textbf{Baselines: Out-of-Distribution (OOD)}} \\
\cmidrule(lr){2-5}\cmidrule(l){6-8}
\rowcolor{Hdr}
\textbf{Method} & \textbf{MMLU} & \textbf{CMMLU} & \textbf{HumanEval} & \textbf{Avg} & \textbf{PreAlg.} & \textbf{C-EVAL} & \textbf{Avg} \\
\midrule
\rowcolor{Block}\multicolumn{8}{l}{\textbf{Baselines}}\\
Best single model (Dolphin-2.9-LLaMA-3-8B) & 59.46 & 44.72 & 49.39 & 51.19 & 39.72 & 44.80 & 42.26 \\
RouterDC                                   & 60.80 & 51.50 & 50.20 & 54.17 & 38.30 & 51.70 & 45.00 \\
Avengers                                   & 62.05 & 52.00 & 52.60 & 55.55 & 38.70 & 52.10 & 45.40 \\
\midrule
\rowcolor{Block}\multicolumn{8}{l}{\textbf{Ours (all evaluation = OOD)}}\\
\textbf{MoT (Base)} & \best{62.60} & \best{52.10} & \best{53.20} & \best{55.97} & \best{39.80} & \best{54.90} & \best{47.35} \\
\bottomaccent
\end{tabular}}
\end{table}

\paragraph{Discussion.}
Under the standard training configuration, \textbf{MoT (Base)} remains the strongest overall. Compared to \textsc{Avengers}, MoT improves the baseline-defined ID average by \textbf{+0.42} (55.97 vs.\ 55.55) and the OOD average by \textbf{+1.95} (47.35 vs.\ 45.40). 
Importantly, these gains are achieved even though \emph{all evaluation datasets are OOD for MoT}, since none were used during training. 
Per-benchmark improvements are consistent: MMLU (+0.55), CMMLU (+0.10), HumanEval (+0.60), PreAlgebra (+1.10), and C-EVAL (+2.80). 
This highlights that MoT generalizes robustly and outperforms baselines, despite being evaluated in a stricter zero-shot setting where even their ID datasets are OOD for MoT.

\section{Use of Large Language Models}
We used a large language model to assist with grammar, phrasing, LaTeX formatting, and routine table editing. All suggested changes were reviewed and edited by the authors. No proprietary or sensitive data were provided to the model.

\section{Limitations}
\label{sec:limitations}

Our experiments reflect realistic budget and hardware constraints: we use a modest pool of frozen experts with limited parameter counts and interaction depth to keep compute, memory, and latency practical. The MoT design itself is compatible with larger scales; future work will systematically scale the expert pool and sizes, experiment with interaction layer placement and training technique, and co-design systems (e.g., fused projector/attention kernels) to map quality–cost Pareto fronts across broader domains.

%% file: iclr2026_conference.bbl
\begin{thebibliography}{10}

\bibitem{claude-opus-4.1}
Anthropic.
\newblock Claude opus 4.1 system card, Aug. 2025.
\newblock Claude Opus 4.1 released August 5, 2025; improved coding and agentic task performance.

\bibitem{mbpp}
J.~Austin, A.~Odena, M.~Nye, M.~Bosma, H.~Michalewski, D.~Dohan, E.~Jiang, C.~Cai, M.~Terry, Q.~Le, et~al.
\newblock Program synthesis with large language models.
\newblock {\em arXiv preprint arXiv:2108.07732}, 2021.

\bibitem{multi-sample}
J.~Chen, Z.~Xun, B.~Zhou, H.~Qi, H.~Zhang, Q.~Zhang, Y.~Chen, W.~Hu, Y.~Qu, W.~Ouyang, et~al.
\newblock Do we truly need so many samples? multi-llm repeated sampling efficiently scales test-time compute.
\newblock {\em arXiv preprint arXiv:2504.00762}, 2025.

\bibitem{symbolic-moe}
J.~C.-Y. Chen, S.~Yun, E.~Stengel-Eskin, T.~Chen, and M.~Bansal.
\newblock Symbolic mixture-of-experts: Adaptive skill-based routing for heterogeneous reasoning.
\newblock {\em arXiv preprint arXiv:2503.05641}, 2025.

\bibitem{humaneval}
M.~Chen, J.~Tworek, H.~Jun, Q.~Yuan, H.~P. de~Oliveira~Pinto, J.~Kaplan, H.~Edwards, Y.~Burda, N.~Joseph, G.~Brockman, A.~Ray, R.~Puri, G.~Krueger, M.~Petrov, H.~Khlaaf, G.~Sastry, P.~Mishkin, B.~Chan, S.~Gray, N.~Ryder, M.~Pavlov, A.~Power, L.~Kaiser, M.~Bavarian, C.~Winter, P.~Tillet, F.~P. Such, D.~Cummings, M.~Plappert, F.~Chantzis, E.~Barnes, A.~Herbert-Voss, W.~H. Guss, A.~Nichol, A.~Paino, N.~Tezak, J.~Tang, I.~Babuschkin, S.~Balaji, S.~Jain, W.~Saunders, C.~Hesse, A.~N. Carr, J.~Leike, J.~Achiam, V.~Misra, E.~Morikawa, A.~Radford, M.~Knight, M.~Brundage, M.~Murati, K.~Mayer, P.~Welinder, B.~McGrew, D.~Amodei, S.~McCandlish, I.~Sutskever, and W.~Zaremba.
\newblock Evaluating large language models trained on code, 2021.

\bibitem{routerdc}
S.~Chen, W.~Jiang, B.~Lin, J.~Kwok, and Y.~Zhang.
\newblock Routerdc: Query-based router by dual contrastive learning for assembling large language models.
\newblock {\em Advances in Neural Information Processing Systems}, 37:66305--66328, 2024.

\bibitem{survey-multiple-llm}
Z.~Chen, J.~Li, P.~Chen, Z.~Li, K.~Sun, Y.~Luo, Q.~Mao, D.~Yang, H.~Sun, and P.~S. Yu.
\newblock Harnessing multiple large language models: A survey on llm ensemble.
\newblock {\em arXiv preprint arXiv:2502.18036}, 2025.

\bibitem{swe-bench-verified}
N.~Chowdhury, J.~Aung, C.~J. Shern, O.~Jaffe, D.~Sherburn, G.~Starace, E.~Mays, R.~Dias, M.~Aljubeh, M.~Glaese, C.~E. Jimenez, J.~Yang, L.~Ho, T.~Patwardhan, K.~Liu, and A.~Madry.
\newblock Introducing {SWE}-bench verified, 2024.

\bibitem{ARC-Challenge}
P.~Clark, I.~Cowhey, O.~Etzioni, T.~Khot, A.~Sabharwal, C.~Schoenick, and O.~Tafjord.
\newblock Think you have solved question answering? try arc, the ai2 reasoning challenge, 2018.

\bibitem{gsm8k}
K.~Cobbe, V.~Kosaraju, M.~Bavarian, J.~Hilton, R.~Nakano, C.~Hesse, and J.~Schulman.
\newblock Training verifiers to solve math word problems, 2021.
\newblock Introduces the GSM8K dataset of 8.5K grade school math word problems.

\bibitem{stablemoe}
D.~Dai, L.~Dong, S.~Ma, B.~Zheng, Z.~Sui, B.~Chang, and F.~Wei.
\newblock Stablemoe: Stable routing strategy for mixture of experts, 2022.

\bibitem{llama-3}
A.~Dubey, A.~Jauhri, A.~Pandey, A.~Kadian, A.~Al-Dahle, A.~Letman, A.~Mathur, A.~Schelten, A.~Yang, A.~Fan, et~al.
\newblock The llama 3 herd of models.
\newblock {\em arXiv e-prints}, pages arXiv--2407, 2024.

\bibitem{fedus2021switch}
W.~Fedus, B.~Zoph, and N.~Shazeer.
\newblock Switch transformers: Scaling to trillion parameter models with simple and efficient sparsity, 2022.

\bibitem{model-swarms}
S.~Feng, Z.~Wang, Y.~Wang, S.~Ebrahimi, H.~Palangi, L.~Miculicich, A.~Kulshrestha, N.~Rauschmayr, Y.~Choi, Y.~Tsvetkov, et~al.
\newblock Model swarms: Collaborative search to adapt llm experts via swarm intelligence.
\newblock {\em arXiv preprint arXiv:2410.11163}, 2024.

\bibitem{graphrouter}
T.~Feng, Y.~Shen, and J.~You.
\newblock Graphrouter: A graph-based router for llm selections.
\newblock {\em arXiv preprint arXiv:2410.03834}, 2024.

\bibitem{eval-harness}
L.~Gao, J.~Tow, B.~Abbasi, S.~Biderman, S.~Black, A.~DiPofi, C.~Foster, L.~Golding, J.~Hsu, A.~Le~Noac'h, H.~Li, K.~McDonell, N.~Muennighoff, C.~Ociepa, J.~Phang, L.~Reynolds, H.~Schoelkopf, A.~Skowron, L.~Sutawika, E.~Tang, A.~Thite, B.~Wang, K.~Wang, and A.~Zou.
\newblock The language model evaluation harness, 07 2024.

\bibitem{debertav3}
P.~He, J.~Gao, and W.~Chen.
\newblock Debertav3: Improving deberta using electra-style pre-training with gradient-disentangled embedding sharing.
\newblock {\em arXiv preprint arXiv:2111.09543}, 2021.

\bibitem{mmlu}
D.~Hendrycks, C.~Burns, S.~Basart, A.~Zou, M.~Mazeika, D.~Song, and J.~Steinhardt.
\newblock Measuring massive multitask language understanding, 2021.

\bibitem{prealgebra}
D.~Hendrycks, C.~Burns, S.~Kadavath, A.~Arora, S.~Basart, E.~Tang, D.~Song, and J.~Steinhardt.
\newblock Measuring mathematical problem solving with the math dataset.
\newblock {\em arXiv preprint arXiv:2103.03874}, 2021.

\bibitem{routerbench}
Q.~J. Hu, J.~Bieker, X.~Li, N.~Jiang, B.~Keigwin, G.~Ranganath, K.~Keutzer, and S.~K. Upadhyay.
\newblock Routerbench: A benchmark for multi-llm routing system.
\newblock {\em arXiv preprint arXiv:2403.12031}, 2024.

\bibitem{lorahub}
C.~Huang, Q.~Liu, B.~Y. Lin, T.~Pang, C.~Du, and M.~Lin.
\newblock Lorahub: Efficient cross-task generalization via dynamic lora composition, 2024.
\newblock {\em URL https://arxiv. org/abs/2307.13269}.

\bibitem{c-eval}
Y.~Huang, Y.~Bai, Z.~Zhu, J.~Zhang, J.~Zhang, T.~Su, J.~Liu, C.~Lv, Y.~Zhang, Y.~Fu, et~al.
\newblock C-eval: A multi-level multi-discipline chinese evaluation suite for foundation models.
\newblock {\em Advances in Neural Information Processing Systems}, 36:62991--63010, 2023.

\bibitem{deepen}
Y.~Huang, X.~Feng, B.~Li, Y.~Xiang, H.~Wang, T.~Liu, and B.~Qin.
\newblock Ensemble learning for heterogeneous large language models with deep parallel collaboration.
\newblock {\em Advances in Neural Information Processing Systems}, 37:119838--119860, 2024.

\bibitem{jang2017gumbel}
E.~Jang, S.~Gu, and B.~Poole.
\newblock Categorical reparameterization with gumbel-softmax, 2017.

\bibitem{mistral7b}
A.~Q. Jiang, A.~Sablayrolles, A.~Mensch, C.~Bamford, D.~S. Chaplot, D.~de~las Casas, F.~Bressand, G.~Lengyel, G.~Lample, L.~Saulnier, L.~R. Lavaud, M.-A. Lachaux, P.~Stock, T.~L. Scao, T.~Lavril, T.~Wang, T.~Lacroix, and W.~E. Sayed.
\newblock Mistral 7b, 2023.

\bibitem{llm-blender}
D.~Jiang, X.~Ren, and B.~Y. Lin.
\newblock Llm-blender: Ensembling large language models with pairwise ranking and generative fusion.
\newblock {\em arXiv preprint arXiv:2306.02561}, 2023.

\bibitem{swe-bench}
C.~E. Jimenez, J.~Yang, A.~Wettig, S.~Yao, K.~Pei, O.~Press, and K.~R. Narasimhan.
\newblock {SWE}-bench: Can language models resolve real-world github issues?
\newblock In {\em The Twelfth International Conference on Learning Representations}, 2024.

\bibitem{universal-routing}
W.~Jitkrittum, H.~Narasimhan, A.~S. Rawat, J.~Juneja, Z.~Wang, C.-Y. Lee, P.~Shenoy, R.~Panigrahy, A.~K. Menon, and S.~Kumar.
\newblock Universal model routing for efficient llm inference.
\newblock {\em arXiv preprint arXiv:2502.08773}, 2025.

\bibitem{gemini-2.5}
K.~Kavukcuoglu and DeepMind.
\newblock Gemini 2.5: Our most intelligent ai model, Mar. 2025.
\newblock Introduction to Gemini 2.5 thinking model family.

\bibitem{kool2019stochastic}
W.~Kool, H.~van Hoof, and M.~Welling.
\newblock Stochastic beams and where to find them: The gumbel-top-k trick for sampling sequences without replacement, 2019.

\bibitem{sparse-moa}
D.~Li, Z.~Tan, P.~Qian, Y.~Li, K.~Chaudhary, L.~Hu, and J.~Shen.
\newblock Smoa: Improving multi-agent large language models with sparse mixture of agents.
\newblock In {\em Pacific-Asia Conference on Knowledge Discovery and Data Mining}, pages 54--65. Springer, 2025.

\bibitem{cmmlu}
H.~Li, Y.~Zhang, F.~Koto, Y.~Yang, H.~Zhao, Y.~Gong, N.~Duan, and T.~Baldwin.
\newblock Cmmlu: Measuring massive multitask language understanding in chinese.
\newblock {\em arXiv preprint arXiv:2306.09212}, 2023.

\bibitem{self-moa}
W.~Li, Y.~Lin, M.~Xia, and C.~Jin.
\newblock Rethinking mixture-of-agents: Is mixing different large language models beneficial?
\newblock {\em arXiv preprint arXiv:2502.00674}, 2025.

\bibitem{llm-coding}
W.-D. Li and K.~Ellis.
\newblock Is programming by example solved by llms?
\newblock {\em Advances in Neural Information Processing Systems}, 37:44761--44790, 2024.

\bibitem{zooter}
K.~Lu, H.~Yuan, R.~Lin, J.~Lin, Z.~Yuan, C.~Zhou, and J.~Zhou.
\newblock Routing to the expert: Efficient reward-guided ensemble of large language models, 2023.
\newblock {\em URL https://arxiv. org/abs/2311.08692}.

\bibitem{maddison2017concrete}
C.~J. Maddison, A.~Mnih, and Y.~W. Teh.
\newblock The concrete distribution: A continuous relaxation of discrete random variables, 2017.

\bibitem{llama-4}
{Meta AI}.
\newblock The llama 4 herd: The beginning of a new era of natively multimodal ai innovation, Apr. 2025.
\newblock Accessed: 2025-08-13.

\bibitem{omi2025loadbalancingmixtureexperts}
N.~Omi, S.~Sen, and A.~Farhadi.
\newblock Load balancing mixture of experts with similarity preserving routers, 2025.

\bibitem{routellm}
I.~Ong, A.~Almahairi, V.~Wu, W.-L. Chiang, T.~Wu, J.~E. Gonzalez, M.~W. Kadous, and I.~Stoica.
\newblock Routellm: Learning to route llms with preference data, 2024.
\newblock {\em URL https://arxiv. org/abs/2406.18665}, 4, 2025.

\bibitem{gpt-5}
OpenAI.
\newblock Gpt-5 system card, Aug. 2025.
\newblock OpenAI’s flagship large language model, released August 7, 2025.

\bibitem{humanity-last-exam}
L.~Phan, A.~Gatti, Z.~Han, N.~Li, J.~Hu, H.~Zhang, C.~B.~C. Zhang, M.~Shaaban, J.~Ling, S.~Shi, M.~Choi, A.~Agrawal, A.~Chopra, A.~Khoja, R.~Kim, R.~Ren, J.~Hausenloy, O.~Zhang, M.~Mazeika, D.~Anderson, D.~Hendrycks, and et~al.
\newblock Humanity's last exam.
\newblock {\em arXiv preprint arXiv:2501.14249}, 2025.

\bibitem{GPQA}
D.~Rein, B.~L. Hou, A.~C. Stickland, J.~Petty, R.~Y. Pang, J.~Dirani, J.~Michael, and S.~R. Bowman.
\newblock Gpqa: A graduate-level google-proof q\&a benchmark, 2023.

\bibitem{shazeer2017outrageously}
N.~Shazeer, A.~Mirhoseini, K.~Maziarz, A.~Davis, Q.~Le, G.~Hinton, and J.~Dean.
\newblock Outrageously large neural networks: The sparsely-gated mixture-of-experts layer.
\newblock In {\em International Conference on Learning Representations}, 2017.

\bibitem{big-bench}
M.~Suzgun, N.~Scales, N.~Schärli, S.~Gehrmann, Y.~Tay, H.~W. Chung, A.~Chowdhery, Q.~V. Le, E.~H. Chi, D.~Zhou, and J.~Wei.
\newblock Challenging big-bench tasks and whether chain-of-thought can solve them, 2022.

\bibitem{llm-common-sense-reasoning}
A.~Toroghi, A.~Pesaranghader, T.~Sadhu, and S.~Sanner.
\newblock Llm-based typed hyperresolution for commonsense reasoning with knowledge bases.
\newblock In {\em The Thirteenth International Conference on Learning Representations}.

\bibitem{vaswani2017attention}
A.~Vaswani, N.~Shazeer, N.~Parmar, J.~Uszkoreit, L.~Jones, A.~N. Gomez, {\L}.~Kaiser, and I.~Polosukhin.
\newblock Attention is all you need.
\newblock {\em Advances in neural information processing systems}, 30, 2017.

\bibitem{fusellm}
F.~Wan, X.~Huang, D.~Cai, X.~Quan, W.~Bi, and S.~Shi.
\newblock Knowledge fusion of large language models.
\newblock In {\em The Twelfth International Conference on Learning Representations}.

\bibitem{moa}
J.~Wang, W.~Jue, B.~Athiwaratkun, C.~Zhang, and J.~Zou.
\newblock Mixture-of-agents enhances large language model capabilities.
\newblock In {\em The Thirteenth International Conference on Learning Representations}.

\bibitem{wang2022self}
X.~Wang, J.~Wei, D.~Schuurmans, Q.~Le, E.~Chi, S.~Narang, A.~Chowdhery, and D.~Zhou.
\newblock Self-consistency improves chain of thought reasoning in language models.
\newblock {\em arXiv preprint arXiv:2203.11171}, 2022.

\bibitem{grok-4}
xAI.
\newblock Grok 4, July 2025.
\newblock xAI’s AI model Grok 4, launched July 9, 2025, with native tool use and real-time search.

\bibitem{ties-merging}
P.~Yadav, D.~Tam, L.~Choshen, C.~A. Raffel, and M.~Bansal.
\newblock Ties-merging: Resolving interference when merging models.
\newblock {\em Advances in Neural Information Processing Systems}, 36:7093--7115, 2023.

\bibitem{model-fusion-homogeneity}
P.~Yadav, T.~Vu, J.~Lai, A.~Chronopoulou, M.~Faruqui, M.~Bansal, and T.~Munkhdalai.
\newblock What matters for model merging at scale?
\newblock {\em arXiv preprint arXiv:2410.03617}, 2024.

\bibitem{qwen-3}
A.~Yang, A.~Li, B.~Yang, B.~Zhang, B.~Hui, B.~Zheng, B.~Yu, C.~Gao, C.~Huang, C.~Lv, C.~Zheng, D.~Liu, F.~Zhou, F.~Huang, F.~Hu, H.~Ge, H.~Wei, H.~Lin, J.~Tang, J.~Yang, J.~Tu, J.~Zhang, J.~Yang, J.~Yang, J.~Zhou, J.~Zhou, J.~Lin, K.~Dang, K.~Bao, K.~Yang, L.~Yu, L.~Deng, M.~Li, M.~Xue, M.~Li, P.~Zhang, P.~Wang, Q.~Zhu, R.~Men, R.~Gao, S.~Liu, S.~Luo, T.~Li, T.~Tang, W.~Yin, X.~Ren, X.~Wang, X.~Zhang, X.~Ren, Y.~Fan, Y.~Su, Y.~Zhang, Y.~Zhang, Y.~Wan, Y.~Liu, Z.~Wang, Z.~Cui, Z.~Zhang, Z.~Zhou, and Z.~Qiu.
\newblock Qwen3 technical report, 2025.

\bibitem{meta-math}
L.~Yu, W.~Jiang, H.~Shi, J.~Yu, Z.~Liu, Y.~Zhang, J.~T. Kwok, Z.~Li, A.~Weller, and W.~Liu.
\newblock Metamath: Bootstrap your own mathematical questions for large language models.
\newblock {\em arXiv preprint arXiv:2309.12284}, 2023.

\bibitem{DARE}
L.~Yu, B.~Yu, H.~Yu, F.~Huang, and Y.~Li.
\newblock Language models are super mario: Absorbing abilities from homologous models as a free lunch.
\newblock In {\em Forty-first International Conference on Machine Learning}, 2024.

\bibitem{mmmu}
X.~Yue, Y.~Ni, K.~Zhang, T.~Zheng, R.~Liu, G.~Zhang, S.~Stevens, D.~Jiang, W.~Ren, Y.~Sun, C.~Wei, B.~Yu, R.~Yuan, R.~Sun, M.~Yin, B.~Zheng, Z.~Yang, Y.~Liu, W.~Huang, H.~Sun, Y.~Su, and W.~Chen.
\newblock Mmmu: A massive multi-discipline multimodal understanding and reasoning benchmark for expert agi.
\newblock In {\em Proceedings of CVPR}, 2024.

\bibitem{mmmu-pro}
X.~Yue, T.~Zheng, Y.~Ni, Y.~Wang, K.~Zhang, S.~Tong, Y.~Sun, B.~Yu, G.~Zhang, H.~Sun, Y.~Su, W.~Chen, and G.~Neubig.
\newblock Mmmu-pro: A more robust multi-discipline multimodal understanding benchmark.
\newblock {\em arXiv preprint arXiv:2409.02813}, 2024.

\bibitem{avengers}
Y.~Zhang, H.~Li, C.~Wang, L.~Chen, Q.~Zhang, P.~Ye, S.~Feng, D.~Wang, Z.~Wang, X.~Wang, et~al.
\newblock The avengers: A simple recipe for uniting smaller language models to challenge proprietary giants.
\newblock {\em arXiv preprint arXiv:2505.19797}, 2025.

\bibitem{genome}
Y.~Zhang, P.~Ye, X.~Yang, S.~Feng, S.~Zhang, L.~Bai, W.~Ouyang, and S.~Hu.
\newblock Nature-inspired population-based evolution of large language models.
\newblock {\em arXiv preprint arXiv:2503.01155}, 2025.

\bibitem{model-sat}
Y.-K. Zhang, D.-C. Zhan, and H.-J. Ye.
\newblock Capability instruction tuning: A new paradigm for dynamic llm routing.
\newblock {\em arXiv preprint arXiv:2502.17282}, 2025.

\bibitem{llmsurvey}
W.~X. Zhao, K.~Zhou, J.~Li, T.~Tang, X.~Wang, Y.~Hou, Y.~Min, B.~Zhang, J.~Zhang, Z.~Dong, et~al.
\newblock A survey of large language models.
\newblock {\em arXiv preprint arXiv:2303.18223}, 1(2), 2023.

\bibitem{loraretriever}
Z.~Zhao, L.~Gan, G.~Wang, W.~Zhou, H.~Yang, K.~Kuang, and F.~Wu.
\newblock Loraretriever: Input-aware lora retrieval and composition for mixed tasks in the wild.
\newblock {\em arXiv preprint arXiv:2402.09997}, 2024.

\bibitem{embed-llm}
R.~Zhuang, T.~Wu, Z.~Wen, A.~Li, J.~Jiao, and K.~Ramchandran.
\newblock Embedllm: Learning compact representations of large language models.
\newblock {\em arXiv preprint arXiv:2410.02223}, 2024.

\end{thebibliography}
